\documentclass[11pt]{article}

\usepackage[final]{acl}

\usepackage{times}
\usepackage{latexsym}

\usepackage[T1]{fontenc}

\usepackage[utf8]{inputenc}

\usepackage{microtype}

\usepackage{inconsolata}

\usepackage{graphicx}

\usepackage[final]{acl}

\usepackage{times}
\usepackage{latexsym}
\usepackage{booktabs}
\usepackage{algorithm}
\usepackage{algpseudocode}
\usepackage{tabularx}
\usepackage{xcolor}
\usepackage{pifont}

\usepackage{titlesec}
\titlespacing*{\section}{0pt}{1.5ex}{0.8ex}
\titlespacing*{\subsection}{0pt}{1.2ex}{0.6ex}
\titlespacing*{\paragraph}{0pt}{0.8ex}{0.5ex}
\renewcommand{\arraystretch}{0.9}
\usepackage{enumitem}
\setlist{nosep}
\renewcommand{\arraystretch}{0.85}
\newcommand{\cmark}{\textcolor{green!60!black}{\ding{51}}} 
\newcommand{\xmark}{\textcolor{red!70!black}{\ding{55}}}   
\newcommand{\pmark}{\textcolor{orange!80!black}{\ding{108}}} 
\newcommand{\hidden}[1]{}
\usepackage[T1]{fontenc}
\usepackage{caption}
\usepackage{enumitem}[shortlabels]
\setlist{nosep, leftmargin=*}
\renewcommand{\arraystretch}{0.95}

\usepackage[utf8]{inputenc}
\usepackage{amsmath}
\usepackage{microtype}
\usepackage{multirow}
\usepackage{inconsolata}

\usepackage{graphicx}
\title{COCORELI: Enforcing Execution Preconditions for Reliable Collaborative Instruction Following \thanks{Code available at: \url{https://github.com/swarnadeep8597/Cocoreli}}}

\author{
Swarnadeep Bhar\thanks{These authors contributed equally to this work.} \\
IRIT \\
\small{\texttt{sbhar8597@gmail.com}}
\And
Omar Naim\footnotemark[2] \\
Université de Toulouse \\
\small{\texttt{omar.naim.docs@gmail.com}}
\And
Eleni Metheniti \\
ANITI \\
\small{\texttt{lenakmeth@gmail.com}}
\And
Bastien Navarri \\
ANITI \\
\small{\texttt{navarri.bastien@gmail.com}}
\AND
Loïc Cabannes \\
ENS Paris-Saclay \\
\small{\texttt{{loic.cabannes@gmail.com}}}
\And
Morteza Ezzabady \\
IRIT \\
\small{\texttt{morteza.ezzabady@irit.fr}}
\And
Nicholas Asher \\
CNRS, IRIT \\
\small{\texttt{nicholas.asher@irit.fr}}
}

\begin{document}
\maketitle
\begin{abstract}
Autonomous agents executing human instructions must operate reliably even when instructions are incomplete. While recent approaches improve detection of missing information, detection alone is insufficient: agents often proceed to execution even after recognizing underspecification, leading to incorrect or unsafe actions.
We identify this failure as arising from a lack of coupling between detection and execution, and propose that reliable behavior requires enforcing missing information as a precondition for action. We instantiate this principle in {\sc Cocoreli}, a modular architecture that represents task structure, tracks missing information, and blocks execution until required details are resolved through targeted clarification. In {\sc Cocoreli}, detection and prevention are structurally coupled: detecting a missing parameter simultaneously blocks execution. We evaluate {\sc Cocoreli} in a controlled construction environment isolating underspecification and sequential execution. 
{
{\sc Cocoreli} makes execution reliable by imposing explicit task structure: instructions are represented as parameterized executable objects and execution under unresolved specifications is blocked by construction, eliminating hallucinated actions. In contrast, chain-of-thought, prompt-chaining, and ReAct-style reasoning may still execute under incomplete specifications despite high detection rates. The same representation supports abstraction and reuse, and generalizes to API workflow tasks on ToolBench.
}
These results show that reliable collaborative execution {under underspecified instructions }requires architectural enforcement, not just model capability.

\end{abstract}

\section{Introduction}
Autonomous agents increasingly execute actions based on human instructions in collaborative settings such as robotics, manufacturing assistance, and tool-use systems. In these environments, instructions are often incomplete: humans rely on shared context and incremental communication, leaving essential task parameters unspecified. Existing systems frequently proceed by inferring missing details rather than acquiring them through interaction, producing plausible but incorrect actions. In collaborative construction tasks, this can lead to placing components in unintended locations or using incorrect parts. Such failures arise not from linguistic misunderstanding alone, but from acting without sufficient information to execute the task reliably.

Rather than attempting to model the full complexity of human dialogue, we focus on a minimal but necessary component of collaboration: resolving underspecified instructions that lack the information required for correct execution. Underspecification provides a controlled entry point into collaborative interaction because it can be systematically manipulated and directly determines whether an action can be carried out successfully. By isolating this factor, we study how agents detect missing information, request targeted clarifications, and incorporate the acquired information into subsequent actions, without addressing broader conversational phenomena such as negotiation or social reasoning.

Traditional single-shot prompting approaches, including chain-of-thought reasoning, rely on implicit inference to fill missing parameters, often producing confident but incorrect outputs. Recent agentic approaches attempt to mitigate these issues through prompt decomposition, tool use, and interaction loops, yet typically assume that instructions are executable as given or that missing details can be inferred from background knowledge. Without explicit mechanisms to represent task incompleteness and defer execution until required information is obtained, these systems may still act under insufficient specification, reflecting a decoupling between detecting missing information and preventing execution. Increasing model scale does not fundamentally address this limitation, as larger models tend to generate more plausible guesses rather than reliably detecting and resolving uncertainty.

In this paper, we investigate whether reliability under underspecification depends not only on model capability but also on how task structure and uncertainty are represented during execution. 

In many collaborative scenarios, agents must operate in domains where task-specific training data is scarce or where new objects and procedures are introduced dynamically. Retraining or fine-tuning dialogue models for each new task is often impractical. Consequently, many approaches emphasize prompting-based methods and modular architectures that enable agents to interpret instructions and execute actions at runtime without additional training. This motivates designs that explicitly structure interaction and execution rather than depending solely on learned behavior.

To this end, we propose \textbf{{\sc Cocoreli}},
a modular architecture for collaborative task execution under partial, evolving, and incrementally specified instructions. {\sc Cocoreli} represents instructions as structured executable objects whose parameters may be known or missing. Missing elements trigger targeted clarification requests, and execution is deferred until sufficient information is available. The system maintains task-relevant state in external memory and supports reuse of previously constructed structures via abstraction, enabling reconstruction of earlier configurations in new contexts.

We evaluate {\sc Cocoreli} in a controlled collaborative construction environment designed to isolate key conditions that commonly arise in real collaborative tasks: incomplete task specification, evolving state with limited history, and reconstruction of previously observed structures. Controlled tasks allow systematic comparison of different architectural paradigms under well-defined uncertainty, rather than attributing performance differences to confounding factors present in natural datasets. To assess generality beyond spatial construction, we additionally evaluate the abstraction mechanism on ToolBench API tasks, which involve interpreting instructions into structured parameterized actions in a different domain.

\paragraph{Hypothesis.}
Reliable execution under underspecification requires not just detection of missing information, but architectural enforcement that prevents action until missing information is resolved. We test this by asking whether explicit structural coupling of detection and prevention improves reliability over approaches that rely on implicit inference, regardless of model scale. We do not claim {\sc Cocoreli} as a definitive solution; rather, we use it as a minimal instantiation of this property to show that prompting-based paradigms systematically lack it.

We summarize our contributions as follows:

\begin{enumerate}
    \item \textbf{A structured mechanism for execution under underspecification.}
    We introduce {\sc Cocoreli}, an architecture that represents instructions as executable structures with explicit tracking of missing parameters, enabling targeted clarification and deferral of action until sufficient information is available.

    \item \textbf{A controlled evaluation of collaborative execution.}
    We design a collaborative construction environment that isolates underspecification, sequential state changes, and structural reuse, allowing systematic comparison of different execution paradigms under well-defined uncertainty.

    \item \textbf{Empirical analysis of execution reliability.}
    We compare structured parsing, chain-of-thought prompting, and prompt-chaining pipelines, showing that approaches lacking explicit mechanisms for managing incomplete specifications frequently produce incorrect actions.

    \item \textbf{Evidence for structural reuse via abstraction.}
    We demonstrate that the proposed approach supports reconstruction of previously built structures in new contexts and transfers to ToolBench API tasks, suggesting applicability beyond spatial construction.
\end{enumerate}
\begin{table*}[t]
\centering
\footnotesize
\setlength{\tabcolsep}{4pt}
\renewcommand{\arraystretch}{0.9}
\begin{tabular}{lccccccc}
\toprule
\textbf{System} &
\textbf{Mem.} &
\textbf{Grnding} &
\textbf{G.Rails} &
\textbf{Exec} &
\textbf{CQ} &
\textbf{Interact} &
\textbf{P\&P} \\
\midrule
ReAct & \xmark & \cmark & \pmark & \cmark & \xmark & \xmark & \xmark \\
ART & \xmark & \cmark & \pmark & \cmark & \xmark & \xmark & \cmark \\
SayCan & \xmark & \cmark & \cmark & \cmark & \xmark & \xmark & \pmark \\
MapAgent & \xmark & \cmark & \cmark & \cmark & \xmark & \xmark & \cmark \\
Voyager & \cmark & \cmark & \cmark & \cmark & \xmark & \xmark & \cmark \\
ToolFormer & \xmark & \cmark & \xmark & \cmark & \xmark & \xmark & \cmark \\
ToolAlpaca & \xmark & \cmark & \xmark & \cmark & \xmark & \xmark & \cmark \\
PAL & \xmark & \cmark & \cmark & \cmark & \xmark & \xmark & \cmark \\
ClarifyCoder & \xmark & \xmark & \xmark & \xmark & \cmark & \cmark & \pmark \\
ASK-TO-ACT & \xmark & \cmark & \cmark & \cmark & \cmark & \cmark & \pmark \\
LEGOMem & \cmark & \cmark & \pmark & \cmark & \xmark & \xmark & \cmark \\
CoALA (framework) & \cmark & \cmark & \pmark & \cmark & \pmark & \cmark & \cmark \\
COCORELI & \cmark & \cmark & \cmark & \cmark & \cmark & \cmark & \cmark \\
\bottomrule
\end{tabular}
\caption{Positioning COCORELI and prior \emph{task-specific agent systems} within the CoALA perspective.
All systems, including COCORELI, are instantiated in particular environments or domains.
COCORELI differs in that it simultaneously operationalizes all dimensions in a single environment
designed to stress-test grounding, clarification, abstraction, and execution.
\cmark: supported, \xmark: not supported, \pmark: partially or implicitly supported.}
\label{tab:coala_positioning}
\end{table*}

{
\section{Related Work}
\label{sec:rel}

Collaborative instruction following under partial and evolving specifications requires agents to ground language in an external environment, maintain state across interactions, and resolve missing information before executing actions. Prior work has addressed subsets of these challenges through dialogue management, clarification, tool use, and agentic architectures. We focus on approaches most relevant to execution under underspecification and their assumptions about when actions may occur.

\paragraph{Task-oriented dialogue systems.}
Classical task-oriented dialogue systems address incomplete instructions through slot filling and clarification questions within predefined domains. Early collaborative dialogue systems emphasized explicit representations of task state and shared information. TRAINS \citep{ALLEN01011995} modeled dialogue as collaborative plan construction, while the Information State framework \citep{trindi} formalized dialogue management as updates to structured representations of beliefs, goals, and dialogue context. Subsequent work on belief-state tracking and dialogue state tracking (DST) \citep{pomdp-young,jacqmin-etal-2022-follow} established explicit maintenance of task-relevant information and uncertainty as a central component of task-oriented dialogue. More recent schema-guided approaches \citep{rastogi2020schema} represent tasks through explicit intents and slots, enabling generalization across services while preserving structured state tracking. Related work on frame-based common-ground representations \citep{common-ground} similarly models dialogue through partially instantiated structures and uses clarification questions to resolve missing information.

\paragraph{Clarification in interactive systems.}
Recent work explores clarification question generation to improve task performance or recover from errors, including ASK-TO-ACT \citep{zhang2025asktoactenhancingllmstool} and ClarifyCoder \citep{wu2025codelanguagemodelslearn}. In these systems, clarification typically functions as a refinement mechanism within a reasoning pipeline, helping models improve task understanding before continuing execution. While effective for reducing ambiguity, clarification remains primarily a reasoning aid rather than an explicit control mechanism governing whether execution may proceed.

\paragraph{Agentic frameworks and reasoning agents.}
Recent agentic frameworks decompose tasks into sequences of reasoning and action. COALA \citep{sumers:etal:2023} characterizes agents in terms of memory, actions, and decision procedures, while agentic workflow approaches distribute tasks across specialized sub-agents \citep{sudarshan2024agentic}. Related LLM-based systems such as ReAct \citep{yao2023reactsynergizingreasoningacting}, Reflexion \citep{shinn2023reflexionlanguageagentsverbal}, ART \citep{paranjape2023artautomaticmultistepreasoning}, reinforcement-learning-based approaches \citep{ahn2022icanisay}, debate-style reasoning \citep{lin2023swiftsagegenerativeagentfast}, and memory-augmented planning \citep{kong2025mapagenttrajectoryconstructedmemoryaugmentedplanning} improve decision making through iterative reasoning, tool use, or self-correction. These approaches generally assume that instructions are sufficiently specified for execution or rely on post hoc recovery when errors occur.

\paragraph{Interactive and memory-augmented agents.}
Several systems incorporate interaction or memory to improve grounded instruction following. HELPER \citep{sarch2023openendedinstructableembodiedagents} retrieves prior language--program pairs to support future tasks, while other approaches incorporate corrective feedback after execution \citep{mehta2024improvinggroundedlanguageunderstanding}. Such systems improve adaptation and task performance through interaction and experience, but clarification or correction typically occurs after an action has already been proposed or executed.

\paragraph{Collaborative construction and abstraction.}
Collaborative construction and embodied task-learning environments highlight the challenges of translating language into executable action sequences \citep{kanitscheider2021multitaskcurriculumlearningcomplex,lin:etal:2021}. Voyager \citep{wang2023voyageropenendedembodiedagent} demonstrates that reusable skills can emerge through iterative interaction and memory, but stores learned procedures in relatively task-specific forms. More broadly, these systems emphasize planning, execution, and reuse in grounded environments without explicitly addressing underspecification as a first-class architectural concern.

\paragraph{Architectural capability comparison.}
Table~\ref{tab:coala_positioning} summarizes differences among these approaches across dimensions relevant to collaborative execution under underspecification, including persistent memory, environmental grounding, proactive clarification, execution guardrails, and modular composition. Across these lines of work, prior systems provide mechanisms for state tracking, clarification, memory, planning, or tool use, but typically treat missing-information detection and action execution as separate processes. \textsc{Cocoreli} explicitly couples them: instructions are represented as executable task structures with unresolved parameters, missing information triggers targeted clarification, and execution is blocked until required parameters have been acquired. Our focus is therefore not only on improving clarification or reasoning, but on enforcing execution preconditions for reliable collaborative instruction following.

}

\section{The {\sc Cocobots} Task}
\label{sec:coco}

We build upon \textsc{Cocobots} \cite{paetzel2022conversational}, a collaborative construction environment designed to evaluate language agents on instruction following under underspecification, structured spatial reasoning, and interactive clarification. Patterned after the Minecraft Collaborative Building Task \citep{narayan-chen-etal-2019-collaborative}, \textsc{Cocobots} departs from Minecraft-style environments through novel object types, altered interaction rules, and stricter physical constraints. Tasks are programmatically generated from a predefined inventory of parts and placement constraints, enabling controlled manipulation of instruction completeness and spatial complexity.

\paragraph{Task structure.}
As in the Minecraft task, \textsc{Cocobots} involves two agents with asymmetric information: an \textit{Architect}, who has access to a target structure, and a \textit{Builder}, who constructs the structure by following the Architect’s natural language instructions. The Builder does not have access to the target structure except through the instructions issued during interactive conversation and the evolving scene state. Instructions may omit details, requiring the Builder to detect and resolve underspecification through interaction.

\paragraph{Cocobots and objects.}
The task is defined over a discrete $16 \times 16 \times 16$ three-dimensional grid, increasing spatial complexity relative to prior collaborative construction settings. Unlike Minecraft’s uniform cubic blocks, \textsc{Cocobots} features nine synthetic part types with heterogeneous spatial properties, including single-cell components (e.g., screws, nuts, washers) and multi-cell components (e.g., bridges spanning two adjacent cells). Valid placement requires reasoning about orientation, spatial extent, and partial coordinate specifications.

\paragraph{Physical and execution constraints.}
\textsc{Cocobots} enforces strict physical constraints inspired by real-world assembly, including gravity and support dependencies. Incorrect placements are not permitted and cannot be corrected post hoc. As a result, agents must resolve ambiguities and ensure that instructions are sufficiently specified before executing actions.

\paragraph{Instruction characteristics.}
Architect instructions vary in specificity and form, ranging from fully specified commands to underspecified or relational descriptions (e.g., relative position, orientation, or part attributes). Instructions may also introduce previously unseen composite structures, requiring the Builder to interpret descriptions relative to the evolving scene state and to request clarification when required information is missing.



\paragraph{Design rationale.}
\textsc{Cocobots} addresses key limitations of existing collaborative benchmarks: (i) assumptions of complete specifications or permissive error correction, (ii) simplified spatial reasoning with uniform components, and (iii) limited evaluation of abstraction and reuse. Tasks are defined independently of any particular architecture and require correct execution given only the available information. 

By combining underspecified language, structured spatial constraints, and irreversible execution errors, \textsc{Cocobots} provides a controlled testbed for studying collaborative instruction following.

Figure~\ref{fig:illustration} contrasts \textsc{Cocobots} with Minecraft-based tasks. While retaining an intuitive spatial setting, it introduces sufficient novelty and constraint to require genuine interaction, abstraction, and clarification.

\section{{\sc Cocoreli}: A Structured Agentic Architecture}
\label{sec:vision}

\begin{figure}[t]
\centering
\includegraphics[width=\columnwidth]{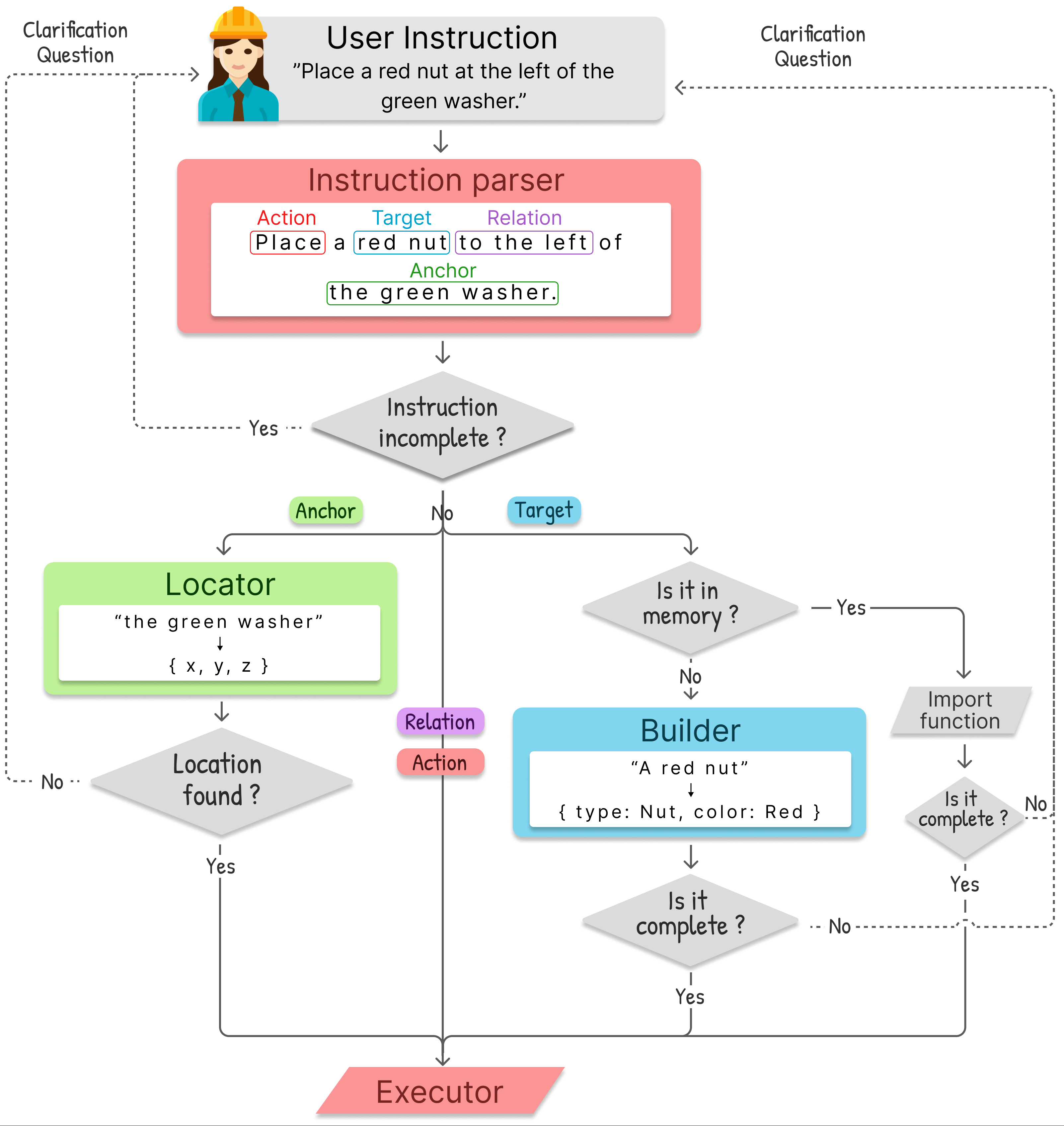}
\vspace{-4pt}
\caption{{\sc Cocoreli}, Our Structured Agentic Architecture}
\vspace{-6pt}
\label{fig:vision}
\end{figure}

{\sc Cocoreli}  is a modular agentic architecture designed for collaborative instruction following under underspecified natural language commands. The architecture addresses three challenges highlighted by the {\sc Cocobots} benchmark: (i) detecting missing or ambiguous information before execution, (ii) grounding relational language in structured spatial representations, and (iii) learning compositional abstractions that generalize across instances. Figure~\ref{fig:vision} provides an overview.

\paragraph{Architectural principles.}
Early experiments showed that monolithic prompting with a single LLM is brittle: long prompts reduce reliability and models often infer missing attributes instead of requesting clarification. {\sc Cocoreli} therefore decomposes the task into six specialized components: \textbf{Instruction Parser}, \textbf{Locator}, \textbf{Builder}, \textbf{External Memory}, \textbf{Discourse Module}, and \textbf{Executor}. Each component uses the same underlying LLM (LLaMA-3.1 8B; \citealp{grattafiori2024llama3herdmodels}) but operates with a specialized prompt and structured inputs.

A key design decision is to represent objects and actions as typed JSON structures whose fields are initialized to \texttt{null}. As dialogue proceeds, these fields are filled with information extracted from instructions or clarification responses. Execution is blocked whenever required fields remain unresolved; the system instead invokes the \textbf{Discourse Module} to generate a targeted clarification question. This mechanism prevents implicit guessing and enforces explicit resolution of missing information before execution.

\paragraph{Grounding and execution.}
Grounding proceeds through a sequence of specialized modules. The \textbf{Instruction Parser} first identifies the object types referenced in the instruction and extracts preliminary attributes from the language. This step filters the relevant object schemas, avoiding prompt bloat from passing all possible part representations to subsequent components.

The filtered object representations are then processed by two complementary modules. The \textbf{Builder} resolves object-specific attributes such as orientation or part configuration, while the \textbf{Locator} interprets spatial descriptions and maps relational expressions to candidate coordinates in the environment. Because different objects support different placement strategies (e.g., bridges span ranges of cells while washers occupy single coordinates), the Locator applies object-specific placement rules when interpreting spatial constraints. Missing object attributes are handled by the Builder and missing spatial parameters by the Locator; unresolved fields remain \texttt{null}, triggering clarification.

\paragraph{Abstraction and memory.}
{\sc Cocoreli} includes an optional \textbf{External Memory} that can store previously constructed structures as relational graphs whose nodes represent parts and whose edges encode spatial relations. When a structure is committed for reuse, an abstraction operation separates structural patterns from instance-specific parameters such as color or absolute position. These abstractions allow previously constructed configurations to be reproduced in new contexts by instantiating them with different parameters.

\paragraph{Execution.}
Once object attributes and spatial parameters are fully specified, the \textbf{Executor} converts the completed representation into a concrete JSON action and validates it against environmental constraints before placement. Errors are therefore avoided through clarification and structured grounding rather than corrected after the fact.
\paragraph{Scope.}
Although evaluated in a collaborative construction environment, the central mechanisms of {\sc Cocoreli} where explicit detection of underspecification, clarification-driven interaction, structured grounding, and abstraction are not specific to physical manipulation. {\sc Cocoreli} enforces a coupling between detection and execution: incomplete task representations cannot be executed. This contrasts with prompting-based approaches, where detection and execution remain decoupled and actions may still be produced under uncertainty. 
{
These mechanisms capture a minimal requirement for reliable execution under underspecified instructions: missing task information should be resolved before acting.}

\section{Experiments for {\sc Cocobots} tasks}

\subsection{Tasks and baselines}\label{sec:baselines}

We evaluate {\sc Cocoreli} and the baselines on a set of {\sc Cocobots} tasks designed as \emph{diagnostic probes} of capabilities required for collaborative instruction following. The goal is not to construct a large-scale benchmark but to isolate common failure modes of language agents, particularly the tendency to execute actions despite incomplete instructions and to examine whether structured parsing and explicit task representations mitigate these failures. 

The tasks are arranged in increasing complexity and progressively introduce additional requirements such as sequential state tracking, underspecification handling, and abstraction.

\begin{enumerate}[label=(\roman*), align=left, labelsep=0pt]

\item \textbf{Single fully-specified instructions.}  
A single instruction specifying placement of one part with complete information.

\item \textbf{Instruction sequences.}  
Multiple fully specified instructions describing a sequence of placements.

\item \textbf{Complex structure construction.}  
Instruction sequences describing multi-part shapes that require maintaining spatial relations across steps.

\item \textbf{Underspecified instructions.}  
Instructions in which one or more required parameters are omitted, requiring the system to detect missing information and avoid execution until it is resolved.

\item \textbf{In-context abstraction and reuse.}  
After constructing a complex structure from instructions, the system must store the resulting configuration and reproduce it in a new context (e.g., at a different location or with different attributes). This task probes whether systems can form reusable structural representations from interaction history or whether explicit memory mechanisms are required.

\end{enumerate}

Tasks (i)–(ii) evaluate instruction parsing and sequential execution. Task (iii) introduces compositional spatial reasoning over multi-part structures. Task (iv) directly tests the ability to detect and resolve underspecified instructions through clarification. Task (v) evaluates whether systems can abstract previously constructed structures and reuse them in new contexts. 

Taken together, these tasks allow us to examine how different architectural choices affect execution reliability as task complexity increases. They also function as \emph{task-based ablations}, revealing which capabilities are necessary for each level of difficulty. Additional task details are provided in Appendix~\ref{Abalation}.

\paragraph{Baselines}

We compare three architectural paradigms for instruction execution that differ in how strongly task structure is enforced.

\textbf{Single-LLM prompting (CoT).}
The first baseline follows the monolithic prompting paradigm in which a single language model is responsible for interpreting instructions, reasoning about missing information, and generating the final action. We use chain-of-thought (CoT) prompting to encourage intermediate reasoning before producing the final placement specification. This setup reflects the common end-to-end approach where the entire task is solved within a single prompt without explicit intermediate control.

We evaluate this baseline using Claude-3.5 Sonnet \citep{claude} and GPT-4.1 \citep{OpenAI2025}. For tasks requiring reuse or abstraction, previously issued instructions must be repeated in the prompt since the model has no external memory. Preliminary experiments with LLaMA-3.1-8B showed substantially lower accuracy, so we report results for stronger models in this paradigm.



\textbf{Agentic LLM decomposition.}
The second baseline represents an agentic design in which the task is decomposed into subtasks handled by specialized prompts, while each subtask is still solved end-to-end by an LLM. In this setup, the model executes instructions step-by-step for Tasks (i)–(iv), and for abstraction (Task v) is prompted to translate sequences of placement instructions into a reusable Python function that captures the structure of the instruction sequence. If the generated code is correct, reconstructing the structure reduces to invoking this function with new parameters. 

This baseline therefore introduces modular task decomposition while retaining LLM-driven execution, without explicitly coupling clarification generation with execution control. We implement this approach using LLaMA-3-70B \citep{grattafiori2024llama3herdmodels}.



\textbf{Structured parsing architecture.}
{\sc Cocoreli} represents a third design paradigm in which subtasks are further decomposed into LLM-based interpretation and deterministic execution components. Instead of solving subtasks end-to-end, the LLM primarily produces structured representations that are validated and executed by specialized modules. This allows explicit detection of missing parameters and controlled clarification before execution.

\paragraph{Comparison scope.}
The goal of these baselines is not exhaustive optimization but to represent common architectural paradigms and test whether improved reasoning and detection prevent execution under incomplete specifications. The single-LLM baseline reflects monolithic prompting, the agentic baseline reflects modular decomposition with LLM-driven components, and {\sc Cocoreli} represents structured parsing with explicit task representations. Our comparison therefore examines how increasing levels of structural control affect execution reliability under underspecified instructions.

Further implementation details for the baselines and {\sc Cocoreli} are provided in Appendix~\ref{app:technical-specs}.

\section{Results on {\sc Cocobots} tasks}\label{sec:results}

\paragraph{Results of Tasks (i) and (ii).}
Fully specified single-part instructions were handled reliably by all systems, with only minor errors in coordinate recovery for the CoT baselines. Sequences of two instructions (Task ii) proved more challenging, particularly for the single-LLM baselines, which showed reduced accuracy in identifying the properties of the second object. As discussed further in Task (iv), this difficulty becomes more pronounced when the second instruction is underspecified. Detailed results for Tasks (i) and (ii) are reported in Tables~\ref{tab:task-1} and~\ref{tab:task-2} in Appendix~B.

\paragraph{Constructing complex shapes (Task iii).}
Task (iii) evaluates the ability to follow instruction sequences describing multi-part structures. A strict metric considers a shape correct only if all instructions are executed correctly.
However, because complex structures consist of multiple instructions, a more informative measure is the proportion of correctly executed steps within each structure. Table~\ref{tab:task-3} in Appendix~\ref{app:task-iii} reports this step-level accuracy.

Under this metric, {\sc Cocoreli} achieves the highest overall accuracy (78.57\%), outperforming both the CoT baselines and the agentic baseline. Notably, {\sc Cocoreli} is the only system that partially reconstructs the most complex structure (the “Moroccan bridge”), which requires both abstraction and subsequent reference to the structure (see Task~(v)). None of the other systems were able to execute this correctly. These results suggest that explicit task decomposition and structured execution improve robustness when instructions involve multiple spatial relations.

\begin{table}[t]
\centering
\small
\begin{tabular}{lcc}
\toprule
\textbf{System} & \textbf{Step Acc (\%)} & \textbf{Shapes Fully Built (\%)} \\
\midrule
GPT-4.1 (CoT) & 54.67 & 40 \\
Claude 3.5 (CoT) & 69.04 & 60 \\
Agentic LLM & 69.04 & 70 \\
{\sc Cocoreli} & \textbf{78.57} & 50 \\
\bottomrule
\end{tabular}
\caption{Performance on complex shape construction (Task iii). 
Step accuracy measures the proportion of correctly executed instructions across shapes. 
The full per-shape breakdown is provided in Appendix~\ref{app:task-iii}.}
\label{tab:task-3}
\end{table}
\paragraph{Underspecified instructions (Task iv).}
Task (iv) evaluates the ability to detect and resolve missing information; all metrics are computed over underspecified inputs. We first test underspecified single-part instructions, where systems must detect missing parameters by returning \textit{null} values and, when possible, generate a clarification question (CQ).

As shown in Table~\ref{tab:task-iv-single}, detection performance varies across models, with single-LLM baselines often failing to identify missing parameters. Only the agentic baseline and {\sc Cocoreli} can issue clarification questions. Once the missing information is provided, both systems correctly incorporate it without hallucinations.

We next evaluate underspecified instructions involving two parts (Table~\ref{tab:task-iv-two}). {\sc Cocoreli} reliably detects missing information, asks clarification questions, and updates the parse without hallucinations. In contrast, CoT-style prompting infers missing attributes and hallucinates part properties. ReAct (implemented over GPT-4.1) and the agentic baseline improve detection and generate clarification-like queries, but do not enforce execution constraints, often proceeding under incomplete specifications and leading to persistent hallucinations. This shows that detection alone is insufficient, as baseline systems often infer missing attributes, leading to execution on incorrect but fully specified parses. Additional analysis of user burden is provided in Appendix~\ref{user-burden}.
\begin{table}[t]
    \centering
    \footnotesize
    \renewcommand{\arraystretch}{0.75}
    \setlength{\tabcolsep}{3.3pt}
    \begin{tabular}{@{}lcccc@{}} 
    \toprule
    \textbf{Shape} & \textbf{GPT} & \textbf{Claude} & \textbf{Agentic} & \textbf{{\sc Cocoreli}} \\ \midrule
    \begin{tabular}[c]{@{}l@{}}Detected\end{tabular} & 27.2\hidden{\textbf{100}} & 49.4\hidden{82.8} & \textbf{100} & \textbf{100} \\ \midrule
    CQs Asked & N/A & N/A & \textbf{100} & \textbf{100} \\ \midrule
    \begin{tabular}[c]{@{}l@{}}Correct (CQ)\end{tabular} & N/A & N/A & \textbf{100} & \textbf{100} \\  \midrule
    Hallucinations:& & & & \\
    On Parts & 20.0 & 0.0 & 0.00 & 0.00\\
    On Colors & 2.1 & 2.1 & 34   & 0.00\\
    On Location & 100 & 38.9 &  51.9 & 0.00 \\ \bottomrule
    \end{tabular}
    
    \caption{Accuracy (\%) on Task (iv): underspecified instructions \underline{\smash{describing one part}}. 
    }
    \label{tab:task-iv-single}
\vspace{10pt}
    \hidden{
    \begin{tabular}{@{}lcccc@{}}  
    \toprule
    \textbf{Shape} & \textbf{GPT-4.1} & \textbf{Claude} & \textbf{Agentic} & \textbf{{\sc Cocoreli}} \\ \midrule
    \begin{tabular}[c]{@{}l@{}}Missing Info \\ Detected\end{tabular} & 75.00 & 43.47 & 83.67 & \textbf{100} \\ \midrule
    CQs Asked & N/A & N/A & 83.67 & \textbf{100} \\ \midrule
    \begin{tabular}[c]{@{}l@{}}Correct Parses \\ (After CQ)\end{tabular} & N/A & N/A & \textbf{100} & \textbf{100} \\ \midrule
    Hallucinations& & & & \\
    Parts & 32.69 & 38.46 & 34.21 &0.00\\
    Colors & 27.31 & 56.94 & 19.64 & 0.00\\
    Location & 14.29 & 66.67 & 1.00& 0.00 \\ \bottomrule
    \end{tabular}
    
    \caption{Accuracy (\%) on Task (iv): underspecified instructions \underline{\smash{describing two parts}}. 
    }    \label{tab:task-iv-two}
    }
    \setlength{\tabcolsep}{3pt}

\begin{tabular}{@{}lccccc@{}}
\toprule
\textbf{Shape} & \textbf{GPT} & \textbf{Claude} & \textbf{Agentic} & \textbf{ReAct} & \textbf{{\sc Cocoreli}} \\
\midrule
Detected & 54.5\hidden{75.00} & 53.5 \hidden{43.47} & 98.5 & 96.5\hidden{96.5} & 96.5\hidden{\textbf{100}} \\
CQs Asked & N/A & N/A & 98.5 & 96.5 & \textbf{100} \\
Correct (CQ) & N/A & N/A & \textbf{100} & 96.5\hidden{\textbf{100}} & \textbf{100} \\
\midrule
\multicolumn{6}{l}{Hallucinations:} \\
Parts & 100\hidden{32.69} & 100\hidden{38.46} & 97.5 & 0.5 & 0.00 \\
Colors & 45.5\hidden{27.31} & 46.5\hidden{56.94} & 44.6 & 3.0 & 0.00 \\
Location & 100\hidden{14.29} & 100\hidden{66.67} & 78.2 & 1.5 & 0.00 \\
\bottomrule
\end{tabular}
\caption{Accuracy (\%) on Task (iv): underspecified instructions \underline{\smash{describing two parts}}. Part hallucination is reported for completeness; part identity is often inferable from context.
}
\label{tab:task-iv-two}
\end{table}

\paragraph{In-context abstraction and reuse (Task v).}
Task (v) evaluates whether systems can abstract a structure from instructions and reproduce it in later interactions. Models first construct nine complex structures and are then asked to recreate them without restating the original instructions. A reproduced structure $R$ is considered correct if there exists a bijection $f$ between the parts of $R$ and those of the original structure $O$ that preserves all relative spatial relations.

Table~\ref{tab:task-v} shows that all systems reproduce simple structures successfully. The agentic baseline fails on medium-sized shapes (16–18 parts), while the CoT baselines succeed on these cases but fail on the most complex structures. {\sc Cocoreli} reconstructs all evaluated structures, including the most complex ones.

The differences in performance reflect the nature of the tasks each system must solve. In the CoT and agentic baselines, abstraction requires the model to infer a reusable procedure from a sequence of placement instructions, which involves planning and program synthesis. In particular, the agentic baseline must generate Python code that captures the structure of the instruction sequence, and reconstruction succeeds only if this generated program is correct.
\vspace{-3pt}
In contrast, {\sc Cocoreli} treats abstraction as a structured parsing problem. The system extracts placement information from instructions and represents the resulting configuration as a relational graph with parts as nodes and spatial relations as edges. Reconstructing a structure reduces to re-instantiating this graph with new parameters. The higher accuracy observed for {\sc Cocoreli} thus reflects explicit structural representations rather than superior reasoning ability. See Appendix~\ref{app:guarantee}.
\section{Learning new functions in context in ToolBench}
\begin{table}[ht]
\centering
\small
\setlength{\tabcolsep}{4pt}
\begin{tabular}{lcccc@{\hspace{4pt}}}
\toprule
\textbf{Shape Category} & \textbf{GPT} & \textbf{Claude} & \textbf{Agentic} & \textbf{{\sc Cocoreli}} \\
\midrule
Simple (4 shapes) & 4/4 & 4/4 & 4/4 & 4/4 \\
Medium (3 shapes) & 3/3 & 3/3 & 1/3 & 3/3 \\
Complex (2 shapes) & 0/2 & 0/2 & 0/2 & 2/2 \\
\bottomrule
\end{tabular}
\caption{Summary results for Task (v): abstraction and reproduction of novel shapes. 
Full per-shape results are provided in Appendix~\ref{app:task-v}.}
\label{tab:task-v}
\end{table}
To examine whether the abstraction mechanism of {\sc Cocoreli} transfers
beyond spatial construction, we evaluate it on the ToolBench dataset
\citep{qin2023toolllmfacilitatinglargelanguage}. ToolBench contains workflows
composed of API calls that access external services. While this domain differs
from {\sc Cocobots}, both settings require mapping natural language
instructions to structured, parameterized actions that can later be reused with
new arguments. {
We use ToolBench because it provides workflow demonstrations that can be abstracted and re-instantiated with new parameters, allowing us to directly evaluate structural reuse beyond the spatial domain.}

ToolBench does not contain underspecified instructions and therefore does
not exercise the clarification mechanism. Instead, the experiment isolates the
abstraction capability: given a workflow and new input parameters, the system
must reproduce the correct function calls with updated arguments. Because the
task concerns structured reuse rather than interactive execution, we compare
{\sc Cocoreli} only with single-LLM CoT baselines.

\paragraph{Methodology}
After preprocessing the workflows (Appendix~\ref{sec:toolbench-data}), we
provide the system with (i) the original user instruction and corresponding API
call sequence and (ii) new parameter values. {\sc Cocoreli} parses the workflow
instruction with the Instruction Parser and produces a structured
representation \texttt{workflow\_i}. When given the cue {\em this is a
workflow\_i}, the Builder abstracts this representation and stores the
template in External Memory. The stored template can then be instantiated with
new arguments to produce the updated API calls. In contrast, the CoT baseline
(Claude 3.5 Sonnet) must infer the workflow pattern directly from the prompt
and regenerate the calls with modified parameters.

\paragraph{Results}
On 100 ToolBench workflows, the CoT baseline (Claude) achieved
36.4 precision, 26.9 recall, and 30.9 F1, whereas {\sc Cocoreli}
achieved 100 on all three metrics. Precision measures the
proportion of generated API calls that match the expected
workflow, while recall measures the proportion of expected calls
correctly reproduced.

These results should not be interpreted as demonstrating superior
reasoning. In ToolBench the baseline must infer workflow structure
through planning and pattern induction, whereas {\sc Cocoreli}
converts the task into structured parsing followed by deterministic
abstraction via its relational representation. The experiment
therefore illustrates how explicit structural representations and
external memory enable reliable workflow reuse rather than
indicating that {\sc Cocoreli} performs a harder task.

Overall, reliable agent execution benefits from explicit mechanisms for both clarification of missing task parameters
and reusable task abstractions.
{
\section{Conclusion}

Reliable execution under underspecified instructions requires more than detecting missing information. Agents must also prevent execution until the information necessary for task completion has been acquired. To investigate this principle, we introduced \textsc{Cocoreli}, a modular architecture that represents instructions as executable task structures, explicitly tracks unresolved parameters, and couples clarification with execution control.

Experiments in \textsc{Cocobots} show that enforcing execution preconditions through structured representations and targeted clarification prevents hallucinated actions under incomplete specifications, whereas prompting-based and agentic baselines may still execute despite detecting missing information. We further demonstrated that the same representations support abstraction and reuse beyond spatial construction through transfer to ToolBench workflows.

Taken together, our results suggest that reliable instruction following under incomplete task specifications depends not only on model capability, but also on architectural mechanisms that explicitly represent task requirements and enforce their satisfaction before execution. We hope that \textsc{Cocoreli} and \textsc{Cocobots} provide useful foundations for future work on collaborative agents operating under incomplete information.
}
\section*{Acknowledgments}
This work was supported by the FRAE-RAKEL project (R093–FRAE–RAKEL), the ANR project COCOBOTS (ANR-21-FAI2-0005), and the AI Cluster ANITI (ANR-19-PI3A-0004). The experiments reported in this paper were performed using HPC resources provided by CALMIP (Grants 2016-P23060 and M24047).

\section*{Limitations}
{\sc Cocoreli} focuses on reliability under underspecified instructions rather than providing a complete architecture for collaborative agents. Several limitations remain.


{
First, our approach assumes that tasks can be represented through structured schemas. In our experiments, API functions and object specifications are expressed in JSON, which enables explicit detection of missing parameters. Real-world systems may require additional preprocessing or schema design to convert raw tool interfaces or environment states into such structured representations. Moreover, identifying which recurring patterns in demonstrations or interaction histories should be abstracted into reusable task structures remains a challenging problem in its own right. Our work assumes that such reusable patterns can be identified and focuses instead on their structured representation and subsequent reuse.
}
{
Second, our guarantees apply only when underspecification can be represented within a known task schema. {\sc Cocoreli} does not address cases where instructions are fully specified but incorrectly parsed, where users refer to entities or actions that fall outside the schema, or where ambiguity arises from pragmatic intent rather than missing task parameters. These forms of uncertainty require richer semantic representations, schema adaptation mechanisms, and more sophisticated models of user intent.
}
{
Although \textsc{Cocobots} is intentionally controlled to isolate underspecification, sequential state tracking, and structural reuse, future work should evaluate these principles in richer collaborative environments involving more complex planning and grounding requirements.
}

Third, our conversational component models only a narrow form of dialogue: clarification for resolving missing task parameters. It does not address other collaborative discourse phenomena such as corrections, negotiation of goals, or extended multi-turn reasoning. Handling richer interaction patterns would require more sophisticated dialogue management mechanisms.

An additional limitation concerns the experimental comparison. Our baselines represent three commonly used architectural paradigms rather than matched implementations on a shared backbone. Although our goal is to evaluate the effect of architectural design rather than optimize individual prompting strategies, future work should compare \textsc{Cocoreli} against equally engineered CoT or agentic pipelines using the same underlying language model. Such experiments would further isolate the contribution of the execution architecture from differences in prompt engineering effort and model capabilities.

Finally, {\sc Cocoreli} does not include an explicit planning module for decomposing high-level goals into sequences of actions. In our experiments, instructions are assumed to be provided by an Architect agent, and integrating structured clarification and abstraction mechanisms with planning remains an open challenge.

Despite these limitations, our experiments highlight a necessary architectural capability for reliable agent execution under incomplete task specifications. Systems that lack explicit mechanisms for detecting missing task parameters and representing reusable task structures tend to rely on implicit inference, which frequently leads to incorrect or inconsistent actions. {
Our results therefore suggest that clarification and structured abstraction are important architectural components for reliable execution under incomplete task specifications.} 

\enlargethispage{2\baselineskip}
\bibliography{latex/multimodal, more_citations}

@inproceedings{zhang:etal:2022,
  author={Honghua Zhang and Liunian Harold Li and Tao Meng and Kai-Wei Chang and Guy Van den Broeck},
  title={On the Paradox of Learning to Reason from Data},
  year={2023},
  cdate={1672531200000},
  pages={3365-3373},
  url={https://doi.org/10.24963/ijcai.2023/375},
  booktitle={IJCAI}
}

@inproceedings{gsm-symbolic,
    title={GSM-Symbolic: Understanding the Limitations of Mathematical Reasoning in Large Language Models}, 
      author={Iman Mirzadeh and Keivan Alizadeh and Hooman Shahrokhi and Oncel Tuzel and Samy Bengio and Mehrdad Farajtabar},
      year={2025},
      eprint={2410.05229},
      archivePrefix={arXiv},
      primaryClass={cs.LG},
      url={https://arxiv.org/abs/2410.05229}, 
}

@inproceedings{narayan-chen-etal-2019-collaborative,
    title = "{Collaborative Dialogue in {M}inecraft}",
    author = "Narayan-Chen, Anjali  and
      Jayannavar, Prashant  and
      Hockenmaier, Julia",
    editor = "Korhonen, Anna  and
      Traum, David  and
      M{\`a}rquez, Llu{\'\i}s",
    booktitle = "Proceedings of the 57th Annual Meeting of the Association for Computational Linguistics",
    month = jul,
    year = "2019",
    address = "Florence, Italy",
    publisher = "Association for Computational Linguistics",
    url = "https://aclanthology.org/P19-1537",
    doi = "10.18653/v1/P19-1537",
    pages = "5405--5415",
}

@article{lin:etal:2021,
  title={Truthfulqa: Measuring how models mimic human falsehoods},
  author={Lin, Stephanie and Hilton, Jacob and Evans, Owain},
  journal={arXiv preprint arXiv:2109.07958},
  year={2021}
}

@article{
    wang2023voyageropenendedembodiedagent,
    title={{Voyager: An Open-Ended Embodied Agent with Large Language Models}},
    author={Guanzhi Wang and Yuqi Xie and Yunfan Jiang and Ajay Mandlekar and Chaowei Xiao and Yuke Zhu and Linxi Fan and Anima Anandkumar},
    journal={Transactions on Machine Learning Research},
    issn={2835-8856},
    year={2024},
    url={https://openreview.net/forum?id=ehfRiF0R3a},
    note={}
    }

@misc{ahn2022icanisay,
      title={{Do As I Can, Not As I Say: Grounding Language in Robotic Affordances}}, 
      author={Michael Ahn and Anthony Brohan and Noah Brown and Yevgen Chebotar and Omar Cortes and Byron David and Chelsea Finn and Chuyuan Fu and Keerthana Gopalakrishnan and Karol Hausman and Alex Herzog and Daniel Ho and Jasmine Hsu and Julian Ibarz and Brian Ichter and Alex Irpan and Eric Jang and Rosario Jauregui Ruano and Kyle Jeffrey and Sally Jesmonth and Nikhil J Joshi and Ryan Julian and Dmitry Kalashnikov and Yuheng Kuang and Kuang-Huei Lee and Sergey Levine and Yao Lu and Linda Luu and Carolina Parada and Peter Pastor and Jornell Quiambao and Kanishka Rao and Jarek Rettinghouse and Diego Reyes and Pierre Sermanet and Nicolas Sievers and Clayton Tan and Alexander Toshev and Vincent Vanhoucke and Fei Xia and Ted Xiao and Peng Xu and Sichun Xu and Mengyuan Yan and Andy Zeng},
      year={2022},
      eprint={2204.01691},
      archivePrefix={arXiv},
      primaryClass={cs.RO},
      url={https://arxiv.org/abs/2204.01691}, 
}

@misc{yao2023reactsynergizingreasoningacting,
      title={{ReAct: Synergizing Reasoning and Acting in Language Models}}, 
      author={Shunyu Yao and Jeffrey Zhao and Dian Yu and Nan Du and Izhak Shafran and Karthik Narasimhan and Yuan Cao},
      year={2023},
      eprint={2210.03629},
      archivePrefix={arXiv},
      primaryClass={cs.CL},
      url={https://arxiv.org/abs/2210.03629}, 
}

@inproceedings{yao2023treethoughtsdeliberateproblem,
 author = {Yao, Shunyu and Yu, Dian and Zhao, Jeffrey and Shafran, Izhak and Griffiths, Tom and Cao, Yuan and Narasimhan, Karthik},
 booktitle = {Advances in Neural Information Processing Systems},
 editor = {A. Oh and T. Naumann and A. Globerson and K. Saenko and M. Hardt and S. Levine},
 pages = {11809--11822},
 publisher = {Curran Associates, Inc.},
 title = {{Tree of Thoughts: Deliberate Problem Solving with Large Language Models}},
 url = {https://proceedings.neurips.cc/paper_files/paper/2023/file/271db9922b8d1f4dd7aaef84ed5ac703-Paper-Conference.pdf},
 volume = {36},
 year = {2023}
}

@misc{long2023largelanguagemodelguided,
      title={Large Language Model Guided Tree-of-Thought}, 
      author={Jieyi Long},
      year={2023},
      eprint={2305.08291},
      archivePrefix={arXiv},
      primaryClass={cs.AI},
      url={https://arxiv.org/abs/2305.08291}, 
}

@misc{wang2023selfconsistencyimproveschainthought,
      title={Self-Consistency Improves Chain of Thought Reasoning in Language Models}, 
      author={Xuezhi Wang and Jason Wei and Dale Schuurmans and Quoc Le and Ed Chi and Sharan Narang and Aakanksha Chowdhery and Denny Zhou},
      year={2023},
      eprint={2203.11171},
      archivePrefix={arXiv},
      primaryClass={cs.CL},
      url={https://arxiv.org/abs/2203.11171}, 
}

@inproceedings{naim:asher:2024b,
  title={Re-examining learning linear functions in context},
  author={Naim, Omar and Fouilh{\'e}, Guilhem and Asher, Nicholas},
  booktitle={German Conference on Artificial Intelligence (K{\"u}nstliche Intelligenz)},
  pages={104--117},
  year={2025},
  organization={Springer}
}

@article{naim:asher:2025,
  title={Analyzing limits for in-context learning},
  author={Naim, Omar and Bolte, J{\'e}r{\^o}me and Asher, Nicholas},
  journal={arXiv preprint arXiv:2502.03503},
  year={2025}
}

@misc{ssa,
      title={SSA: Improving Performance With a Better Scoring Function}, 
      author={Omar Naim and Swarnadeep Bhar and Jérôme Bolte and Nicholas Asher},
      year={2026},
      eprint={2508.14685},
      archivePrefix={arXiv},
      primaryClass={cs.CL},
      url={https://arxiv.org/abs/2508.14685}, 
}

@misc{tale,
      title={TELL-TALE: Task Efficient LLMs with Task Aware Layer Elimination}, 
      author={Omar Naim and Krish Sharma and Niyar R Barman and Nicholas Asher},
      year={2026},
      eprint={2510.22767},
      archivePrefix={arXiv},
      primaryClass={cs.LG},
      url={https://arxiv.org/abs/2510.22767}, 
}

@Inbook{Harel1984,
author="Harel, David",
editor="Gabbay, D.
and Guenthner, F.",
title="Dynamic Logic",
bookTitle="Handbook of Philosophical Logic: Volume II: Extensions of Classical Logic",
year="1984",
publisher="Springer Netherlands",
address="Dordrecht",
pages="497--604",
isbn="978-94-009-6259-0",
doi="10.1007/978-94-009-6259-0_10",
url="https://doi.org/10.1007/978-94-009-6259-0_10"
}

@INPROCEEDINGS{1168143,
  author={Jenkins, W. and Mather, B. and Munson, D.},
  booktitle={ICASSP '85. IEEE International Conference on Acoustics, Speech, and Signal Processing}, 
  title={Nearest neighbor and generalized inverse distance interpolation for Fourier domain image reconstruction}, 
  year={1985},
  volume={10},
  number={},
  pages={1069-1072},
  doi={10.1109/ICASSP.1985.1168143}}

@inproceedings{
qin2023toolllmfacilitatinglargelanguage,
title={Tool{LLM}: Facilitating Large Language Models to Master 16000+ Real-world {API}s},
author={Yujia Qin and Shihao Liang and Yining Ye and Kunlun Zhu and Lan Yan and Yaxi Lu and Yankai Lin and Xin Cong and Xiangru Tang and Bill Qian and Sihan Zhao and Lauren Hong and Runchu Tian and Ruobing Xie and Jie Zhou and Mark Gerstein and dahai li and Zhiyuan Liu and Maosong Sun},
booktitle={The Twelfth International Conference on Learning Representations},
year={2024},
url={https://openreview.net/forum?id=dHng2O0Jjr}
}

@article{sumers:etal:2023,
title={Cognitive Architectures for Language Agents},
author={Theodore Sumers and Shunyu Yao and Karthik R Narasimhan and Thomas L. Griffiths},
journal={Transactions on Machine Learning Research},
issn={2835-8856},
year={2024},
url={https://openreview.net/forum?id=1i6ZCvflQJ},
note={Survey Certification, Featured Certification}
}

@misc{sudarshan2024agentic,
      title={{Agentic LLM Workflows for Generating Patient-Friendly Medical Reports}}, 
      author={Malavikha Sudarshan and Sophie Shih and Estella Yee and Alina Yang and John Zou and Cathy Chen and Quan Zhou and Leon Chen and Chinmay Singhal and George Shih},
      year={2024},
      eprint={2408.01112},
      archivePrefix={arXiv},
      primaryClass={cs.MA},
      url={https://arxiv.org/abs/2408.01112}, 
}

@misc{parisi2022talmtoolaugmentedlanguage,
      title={TALM: Tool Augmented Language Models}, 
      author={Aaron Parisi and Yao Zhao and Noah Fiedel},
      year={2022},
      eprint={2205.12255},
      archivePrefix={arXiv},
      primaryClass={cs.CL},
      url={https://arxiv.org/abs/2205.12255}, 
}

@inproceedings{NEURIPS2024_e4c61f57,
 title={Gorilla: Large Language Model Connected with Massive APIs},
    author={Patil, Shishir G. and Zhang, Tianjun and Wang, Xin and Gonzalez, Joseph E.},
    booktitle = {Advances in Neural Information Processing Systems},
    year={2024},
}

@InProceedings{pmlr-v202-gao23f,
  title = 	 {{PAL}: Program-aided Language Models},
  author =       {Gao, Luyu and Madaan, Aman and Zhou, Shuyan and Alon, Uri and Liu, Pengfei and Yang, Yiming and Callan, Jamie and Neubig, Graham},
  booktitle = 	 {Proceedings of the 40th International Conference on Machine Learning},
  pages = 	 {10764--10799},
  year = 	 {2023},
  editor = 	 {Krause, Andreas and Brunskill, Emma and Cho, Kyunghyun and Engelhardt, Barbara and Sabato, Sivan and Scarlett, Jonathan},
  volume = 	 {202},
  series = 	 {Proceedings of Machine Learning Research},
  month = 	 {23--29 Jul},
  publisher =    {PMLR},
  url = 	 {https://proceedings.mlr.press/v202/gao23f.html}
}

@inproceedings{schick2023toolformer,
 author = {Schick, Timo and Dwivedi-Yu, Jane and Dessi, Roberto and Raileanu, Roberta and Lomeli, Maria and Hambro, Eric and Zettlemoyer, Luke and Cancedda, Nicola and Scialom, Thomas},
 booktitle = {Advances in Neural Information Processing Systems},
 editor = {A. Oh and T. Naumann and A. Globerson and K. Saenko and M. Hardt and S. Levine},
 pages = {68539--68551},
 publisher = {Curran Associates, Inc.},
 title = {Toolformer: Language Models Can Teach Themselves to Use Tools},
 url = {https://proceedings.neurips.cc/paper_files/paper/2023/file/d842425e4bf79ba039352da0f658a906-Paper-Conference.pdf},
 volume = {36},
 year = {2023}
}

@misc{zhang2025asktoactenhancingllmstool,
      title={AskToAct: Enhancing LLMs Tool Use via Self-Correcting Clarification}, 
      author={Xuan Zhang and Yongliang Shen and Zhe Zheng and Linjuan Wu and Wenqi Zhang and Yuchen Yan and Qiuying Peng and Jun Wang and Weiming Lu},
      year={2025},
      eprint={2503.01940},
      archivePrefix={arXiv},
      primaryClass={cs.CL},
      url={https://arxiv.org/abs/2503.01940}, 
}

@misc{wu2025codelanguagemodelslearn,
      title={Can Code Language Models Learn Clarification-Seeking Behaviors?}, 
      author={Jie JW Wu and Manav Chaudhary and Davit Abrahamyan and Arhaan Khaku and Anjiang Wei and Fatemeh H. Fard},
      year={2025},
      eprint={2504.16331},
      archivePrefix={arXiv},
      primaryClass={cs.SE},
      url={https://arxiv.org/abs/2504.16331}, 
}

@inproceedings{dong-etal-2024-survey,
    title = "A Survey on In-context Learning",
    author = "Dong, Qingxiu  and
      Li, Lei  and
      Dai, Damai  and
      Zheng, Ce  and
      Ma, Jingyuan  and
      Li, Rui  and
      Xia, Heming  and
      Xu, Jingjing  and
      Wu, Zhiyong  and
      Chang, Baobao  and
      Sun, Xu  and
      Li, Lei  and
      Sui, Zhifang",
    editor = "Al-Onaizan, Yaser  and
      Bansal, Mohit  and
      Chen, Yun-Nung",
    booktitle = "Proceedings of the 2024 Conference on Empirical Methods in Natural Language Processing",
    month = nov,
    year = "2024",
    address = "Miami, Florida, USA",
    publisher = "Association for Computational Linguistics",
    url = "https://aclanthology.org/2024.emnlp-main.64/",
    doi = "10.18653/v1/2024.emnlp-main.64",
    pages = "1107--1128"
}

@inproceedings{press-etal-2023-measuring,
    title = {{Measuring and Narrowing the Compositionality Gap in Language Models}},
    author = "Press, Ofir  and
      Zhang, Muru  and
      Min, Sewon  and
      Schmidt, Ludwig  and
      Smith, Noah  and
      Lewis, Mike",
    editor = "Bouamor, Houda  and
      Pino, Juan  and
      Bali, Kalika",
    booktitle = "Findings of the Association for Computational Linguistics: EMNLP 2023",
    month = dec,
    year = "2023",
    address = "Singapore",
    publisher = "Association for Computational Linguistics",
    url = "https://aclanthology.org/2023.findings-emnlp.378/",
    doi = "10.18653/v1/2023.findings-emnlp.378",
    pages = "5687--5711"
}

@inproceedings{wei2022chain,
 author = {Wei, Jason and Wang, Xuezhi and Schuurmans, Dale and Bosma, Maarten and Ichter, Brian and Xia, Fei and Chi, Ed and Le, Quoc V and Zhou, Denny},
 booktitle = {Advances in Neural Information Processing Systems},
 editor = {S. Koyejo and S. Mohamed and A. Agarwal and D. Belgrave and K. Cho and A. Oh},
 pages = {24824--24837},
 publisher = {Curran Associates, Inc.},
 title = {Chain-of-Thought Prompting Elicits Reasoning in Large Language Models},
 url = {https://proceedings.neurips.cc/paper_files/paper/2022/file/9d5609613524ecf4f15af0f7b31abca4-Paper-Conference.pdf},
 volume = {35},
 year = {2022}
}

@misc{tang2023toolalpacageneralizedtoollearning,
      title={ToolAlpaca: Generalized Tool Learning for Language Models with 3000 Simulated Cases}, 
      author={Qiaoyu Tang and Ziliang Deng and Hongyu Lin and Xianpei Han and Qiao Liang and Boxi Cao and Le Sun},
      year={2023},
      eprint={2306.05301},
      archivePrefix={arXiv},
      primaryClass={cs.CL},
      url={https://arxiv.org/abs/2306.05301}, 
}

@inproceedings{NEURIPS2024_5aee125f,
title={Jax{MARL}: Multi-Agent {RL} Environments in {JAX}},
author={Alexander Rutherford and Benjamin Ellis and Matteo Gallici and Jonathan Cook and Andrei Lupu and Gar{\dh}ar Ingvarsson and Timon Willi and Akbir Khan and Christian Schroeder de Witt and Alexandra Souly and Saptarashmi Bandyopadhyay and Mikayel Samvelyan and Minqi Jiang and Robert Tjarko Lange and Shimon Whiteson and Bruno Lacerda and Nick Hawes and Tim Rockt{\"a}schel and Chris Lu and Jakob Nicolaus Foerster},
booktitle={Second Agent Learning in Open-Endedness Workshop},
year={2023},
url={https://openreview.net/forum?id=BlhQN9Jfpf}
}

@inproceedings{
liu2024agentbench,
title={AgentBench: Evaluating {LLM}s as Agents},
author={Xiao Liu and Hao Yu and Hanchen Zhang and Yifan Xu and Xuanyu Lei and Hanyu Lai and Yu Gu and Hangliang Ding and Kaiwen Men and Kejuan Yang and Shudan Zhang and Xiang Deng and Aohan Zeng and Zhengxiao Du and Chenhui Zhang and Sheng Shen and Tianjun Zhang and Yu Su and Huan Sun and Minlie Huang and Yuxiao Dong and Jie Tang},
booktitle={The Twelfth International Conference on Learning Representations},
year={2024},
url={https://openreview.net/forum?id=zAdUB0aCTQ}
}

@inproceedings{
wu2024smartplay,
title={SmartPlay : A Benchmark for {LLM}s as Intelligent Agents},
author={Yue Wu and Xuan Tang and Tom Mitchell and Yuanzhi Li},
booktitle={The Twelfth International Conference on Learning Representations},
year={2024},
url={https://openreview.net/forum?id=S2oTVrlcp3}
}

@misc{OpenAI2025, 
    title={{Introducing GPT-4.1 in the API}}, 
    url={https://openai.com/index/gpt-4-1/}, 
    journal={Introducing GPT-4.1 in the API},
    author={OpenAI}, year={2025}, month={Apr}, note={[Accessed on 14.05.2025]}
    }

@misc{claude, 
    title={{Claude 3.5 Sonnet}}, 
    url={https://www.anthropic.com/news/claude-3-5-sonnet}, 
    journal={Claude 3.5 Sonnet},
    author={Anthropic}, year={2024}, month={Jun}, 
    note={[Accessed on 14.05.2025]}
    }

@misc{grattafiori2024llama3herdmodels,
      title={{The Llama 3 Herd of Models}}, 
      author={Meta},
      year={2024},
      eprint={2407.21783},
      archivePrefix={arXiv},
      primaryClass={cs.AI},
      url={https://arxiv.org/abs/2407.21783}, 
}

@misc{huang2025fardecisionmakingllmsevaluating,
      title={{How Far Are We on the Decision-Making of LLMs? Evaluating LLMs' Gaming Ability in Multi-Agent Environments}}, 
      author={Jen-tse Huang and Eric John Li and Man Ho Lam and Tian Liang and Wenxuan Wang and Youliang Yuan and Wenxiang Jiao and Xing Wang and Zhaopeng Tu and Michael R. Lyu},
      year={2025},
      eprint={2403.11807},
      archivePrefix={arXiv},
      primaryClass={cs.AI},
      url={https://arxiv.org/abs/2403.11807}, 
}

@inproceedings{gioacchini-etal-2024-agentquest,
    title = {{AgentQuest: A Modular Benchmark Framework to Measure Progress and Improve {LLM} Agents}},
    author = "Gioacchini, Luca  and
      Siracusano, Giuseppe  and
      Sanvito, Davide  and
      Gashteovski, Kiril  and
      Friede, David  and
      Bifulco, Roberto  and
      Lawrence, Carolin",
    editor = "Chang, Kai-Wei  and
      Lee, Annie  and
      Rajani, Nazneen",
    booktitle = "Proceedings of the 2024 Conference of the North American Chapter of the Association for Computational Linguistics: Human Language Technologies (Volume 3: System Demonstrations)",
    month = jun,
    year = "2024",
    address = "Mexico City, Mexico",
    publisher = "Association for Computational Linguistics",
    url = "https://aclanthology.org/2024.naacl-demo.19/",
    doi = "10.18653/v1/2024.naacl-demo.19",
    pages = "185--193"
}

@inproceedings{10.1145/3586183.3606763,
author = {Park, Joon Sung and O'Brien, Joseph and Cai, Carrie Jun and Morris, Meredith Ringel and Liang, Percy and Bernstein, Michael S.},
title = {Generative Agents: Interactive Simulacra of Human Behavior},
year = {2023},
isbn = {9798400701320},
publisher = {Association for Computing Machinery},
address = {New York, NY, USA},
url = {https://doi.org/10.1145/3586183.3606763},
doi = {10.1145/3586183.3606763},
booktitle = {Proceedings of the 36th Annual ACM Symposium on User Interface Software and Technology},
articleno = {2},
numpages = {22},
location = {San Francisco, CA, USA},
series = {UIST '23}
}

@inproceedings{xu-etal-2024-magic,
    title = "{MA}g{IC}: Investigation of Large Language Model Powered Multi-Agent in Cognition, Adaptability, Rationality and Collaboration",
    author = "Xu, Lin  and
      Hu, Zhiyuan  and
      Zhou, Daquan  and
      Ren, Hongyu  and
      Dong, Zhen  and
      Keutzer, Kurt  and
      Ng, See-Kiong  and
      Feng, Jiashi",
    editor = "Al-Onaizan, Yaser  and
      Bansal, Mohit  and
      Chen, Yun-Nung",
    booktitle = "Proceedings of the 2024 Conference on Empirical Methods in Natural Language Processing",
    month = nov,
    year = "2024",
    address = "Miami, Florida, USA",
    publisher = "Association for Computational Linguistics",
    url = "https://aclanthology.org/2024.emnlp-main.416/",
    doi = "10.18653/v1/2024.emnlp-main.416",
    pages = "7315--7332"
}

@misc{kanitscheider2021multitaskcurriculumlearningcomplex,
      title={Multi-task curriculum learning in a complex, visual, hard-exploration domain: Minecraft}, 
      author={Ingmar Kanitscheider and Joost Huizinga and David Farhi and William Hebgen Guss and Brandon Houghton and Raul Sampedro and Peter Zhokhov and Bowen Baker and Adrien Ecoffet and Jie Tang and Oleg Klimov and Jeff Clune},
      year={2021},
      eprint={2106.14876},
      archivePrefix={arXiv},
      primaryClass={cs.LG},
      url={https://arxiv.org/abs/2106.14876}, 
}

@misc{shinn2023reflexionlanguageagentsverbal,
      title={Reflexion: Language Agents with Verbal Reinforcement Learning}, 
      author={Noah Shinn and Federico Cassano and Edward Berman and Ashwin Gopinath and Karthik Narasimhan and Shunyu Yao},
      year={2023},
      eprint={2303.11366},
      archivePrefix={arXiv},
      primaryClass={cs.AI},
      url={https://arxiv.org/abs/2303.11366}, 
}

@misc{paranjape2023artautomaticmultistepreasoning,
      title={ART: Automatic multi-step reasoning and tool-use for large language models}, 
      author={Bhargavi Paranjape and Scott Lundberg and Sameer Singh and Hannaneh Hajishirzi and Luke Zettlemoyer and Marco Tulio Ribeiro},
      year={2023},
      eprint={2303.09014},
      archivePrefix={arXiv},
      primaryClass={cs.CL},
      url={https://arxiv.org/abs/2303.09014}, 
}

@misc{lin2023swiftsagegenerativeagentfast,
      title={SwiftSage: A Generative Agent with Fast and Slow Thinking for Complex Interactive Tasks}, 
      author={Bill Yuchen Lin and Yicheng Fu and Karina Yang and Faeze Brahman and Shiyu Huang and Chandra Bhagavatula and Prithviraj Ammanabrolu and Yejin Choi and Xiang Ren},
      year={2023},
      eprint={2305.17390},
      archivePrefix={arXiv},
      primaryClass={cs.CL},
      url={https://arxiv.org/abs/2305.17390}, 
}

@misc{kong2025mapagenttrajectoryconstructedmemoryaugmentedplanning,
      title={MapAgent: Trajectory-Constructed Memory-Augmented Planning for Mobile Task Automation}, 
      author={Yi Kong and Dianxi Shi and Guoli Yang and Zhang ke-di and Chenlin Huang and Xiaopeng Li and Songchang Jin},
      year={2025},
      eprint={2507.21953},
      archivePrefix={arXiv},
      primaryClass={cs.HC},
      url={https://arxiv.org/abs/2507.21953}, 
}

@misc{mehta2024improvinggroundedlanguageunderstanding,
      title={Improving Grounded Language Understanding in a Collaborative Environment by Interacting with Agents Through Help Feedback}, 
      author={Nikhil Mehta and Milagro Teruel and Patricio Figueroa Sanz and Xin Deng and Ahmed Hassan Awadallah and Julia Kiseleva},
      year={2024},
      eprint={2304.10750},
      archivePrefix={arXiv},
      primaryClass={cs.CL},
      url={https://arxiv.org/abs/2304.10750}, 
}

@misc{sarch2023openendedinstructableembodiedagents,
      title={Open-Ended Instructable Embodied Agents with Memory-Augmented Large Language Models}, 
      author={Gabriel Sarch and Yue Wu and Michael J. Tarr and Katerina Fragkiadaki},
      year={2023},
      eprint={2310.15127},
      archivePrefix={arXiv},
      primaryClass={cs.AI},
      url={https://arxiv.org/abs/2310.15127}, 
}

@article{ALLEN01011995,
author = {JAMES F. ALLEN and LENHART K. SCHUBERT and GEORGE FERGUSON and PETER HEEMAN and CHUNG HEE HWANG and TSUNEAKI KATO and MARC LIGHT and NATHANIEL MARTIN and BRADFORD MILLER and MASSIMO POESIO and DAVID R. TRAUM},
title = {The TRAINS project: a case study in building a conversational planning agent},
journal = {Journal of Experimental \& Theoretical Artificial Intelligence},
volume = {7},
number = {1},
pages = {7--48},
year = {1995},
publisher = {Taylor \& Francis},
doi = {10.1080/09528139508953799},
URL = {https://doi.org/10.1080/09528139508953799},
eprint = {https://doi.org/10.1080/09528139508953799}
}

@article{trindi,
author = {Larsson, Staffan and Traum, David R.},
title = {Information state and dialogue management in the TRINDI dialogue move engine toolkit},
year = {2000},
issue_date = {September 2000},
publisher = {Cambridge University Press},
address = {USA},
volume = {6},
number = {3–4},
issn = {1351-3249},
url = {https://doi.org/10.1017/S1351324900002539},
doi = {10.1017/S1351324900002539},
journal = {Nat. Lang. Eng.},
month = sep,
pages = {323–340},
numpages = {18}
}

@ARTICLE{pomdp-young,
  author={Young, Steve and Gašić, Milica and Thomson, Blaise and Williams, Jason D.},
  journal={Proceedings of the IEEE}, 
  title={POMDP-Based Statistical Spoken Dialog Systems: A Review}, 
  year={2013},
  volume={101},
  number={5},
  pages={1160-1179},
  doi={10.1109/JPROC.2012.2225812}
  }

@article{rastogi2020schema,
title	= {Towards Scalable Multi-domain Conversational Agents: The Schema-Guided Dialogue Dataset},
author	= {Abhinav Rastogi and Xiaoxue Zang and Srinivas Kumar Sunkara and Raghav Gupta and Pranav Khaitan},
year	= {2019},
journal	= {arXiv preprint arXiv:1909.05855}
}

@inproceedings{common-ground,
author = {Blache, Philippe and Houlès, Matthis},
year = {2021},
month = {11},
pages = {33-43},
title = {Common Ground, Frames and Slots for Comprehension in Dialogue Systems},
doi = {10.5121/csit.2021.112003}
}

@inproceedings{jacqmin-etal-2022-follow,
    title = "``Do you follow me?'': A Survey of Recent Approaches in Dialogue State Tracking",
    author = "Jacqmin, L{\'e}o  and
      Rojas Barahona, Lina M.  and
      Favre, Benoit",
    editor = "Lemon, Oliver  and
      Hakkani-Tur, Dilek  and
      Li, Junyi Jessy  and
      Ashrafzadeh, Arash  and
      Garcia, Daniel Hern{\'a}ndez  and
      Alikhani, Malihe  and
      Vandyke, David  and
      Du{\v{s}}ek, Ond{\v{r}}ej",
    booktitle = "Proceedings of the 23rd Annual Meeting of the Special Interest Group on Discourse and Dialogue",
    month = sep,
    year = "2022",
    address = "Edinburgh, UK",
    publisher = "Association for Computational Linguistics",
    url = "https://aclanthology.org/2022.sigdial-1.33/",
    doi = "10.18653/v1/2022.sigdial-1.33",
    pages = "336--350"
}

@inproceedings{paetzel2022conversational,
    title={Conversational Programming for Collaborative Robots},
    author={Paetzel-Pr{\"u}smann, Maike and Hunter, Julie and Chalamalasetti, Kranti and Thompson, Kate and Nicolaou, Alexandros and G{\"u}ng{\"o}r, Ozan and Schlangen, David and Asher, Nicholas},
    booktitle={ICRA Workshop on Collaborative Robots and Work of the Future (ICRA 2022 CoR-WotF)},
    pages={1--5},
    year={2022}
}

\newpage
\appendix
\section*{Appendices}
\section{Technical specifications}\label{app:technical-specs}
Baseline single LLM models from commercial APIs:
\begin{enumerate}
    \item Claude-3.5-Sonnet (175 billion parameters) API from the official API:  \url{https://claude.ai/new}
    \item GPT-4.1 (unknown parameter size) from the official API: \url{https://platform.openai.com/docs/models/gpt-4.1}
\end{enumerate}

\noindent For the Agentic Model, we used Llama-3.3-70b-Instruct, through the AI/ML API service:  \url{https://aimlapi.com/app/}.

\noindent For the design and implementation of {\sc Cocoreli}, we used LLaMA-3.1 8b from the Ollama library (\url{https://ollama.com}), and from the \texttt{transformers} library from HuggingFace (\url{https://huggingface.co/}). All 8b experiments were done locally without using any commercial APIs. To run the Hugging Face models locally, we used Volta GPUs from NVIDIA.

\section{On the Nature of the Architectural Contribution}
\label{app:guarantee}

The empirical comparisons in this paper may be interpreted as performance comparisons across systems of different scales. We clarify that the primary contribution is architectural rather than performance-driven.

The central claim is not that \textsc{Cocoreli} outperforms larger models, but that it enforces a structural coupling between detecting missing information and executing actions. In \textsc{Cocoreli}, incomplete task representations cannot be executed: if required parameters remain unresolved, execution is blocked. As a result, the absence of hallucinated actions under underspecification follows from this execution model, rather than from improved detection or reasoning capability.

The baseline comparisons are intended to illustrate that this property is absent in prompting-based paradigms. In such systems, detection of missing information and action generation are decoupled processes: even when missing information is identified during reasoning, actions may still be produced based on inferred values. This behavior reflects architectural design choices rather than limitations of model capability.

The difference in model scale between \textsc{Cocoreli} (LLaMA-3.1 8B) and the CoT baselines (GPT-4.1, Claude-3.5-Sonnet) should therefore be interpreted in this context. The structural enforcement of execution preconditions in \textsc{Cocoreli} operates independently of model scale, while prompting-based approaches rely on model behavior to avoid incorrect execution. Such architectures could also benefit from complementary efficiency methods such as TALE, which improves task-specific performance while further reducing model size \citep{tale}.

We note that this guarantee applies to missing-parameter cases. Incorrect but fully specified representations remain possible and represent an important limitation of the current approach.

\section{COCOBOTS-MINECRAFT}
\begin{figure}[t]
    \centering
    \begin{minipage}[t]{0.24\textwidth}
        \centering
        \includegraphics[height=2.05cm]{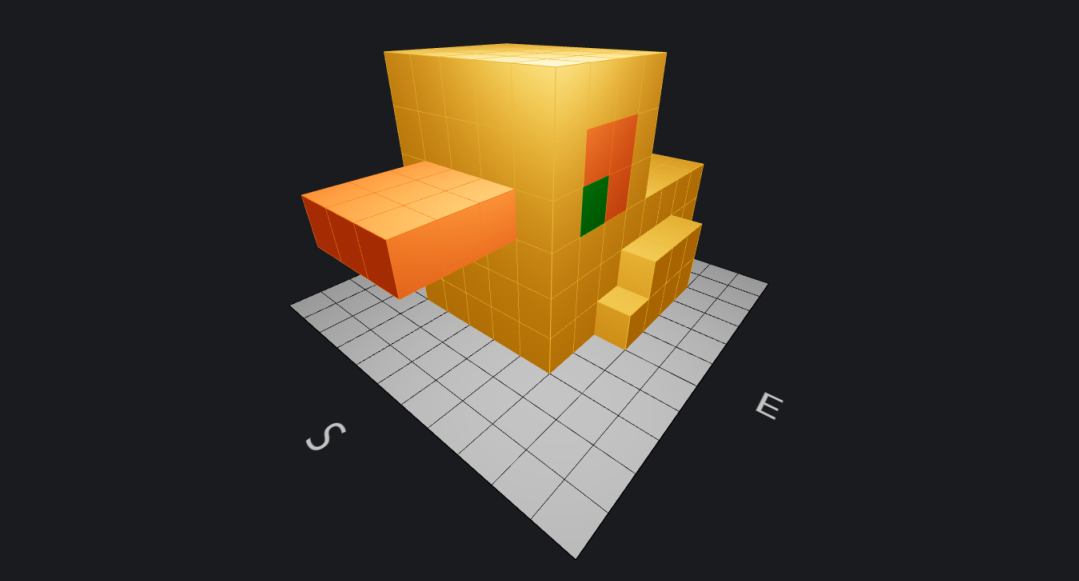}
        Minecraft
    \end{minipage}\hfill
    \begin{minipage}[t]{0.24\textwidth}
        \centering
        \includegraphics[height=2.05cm]{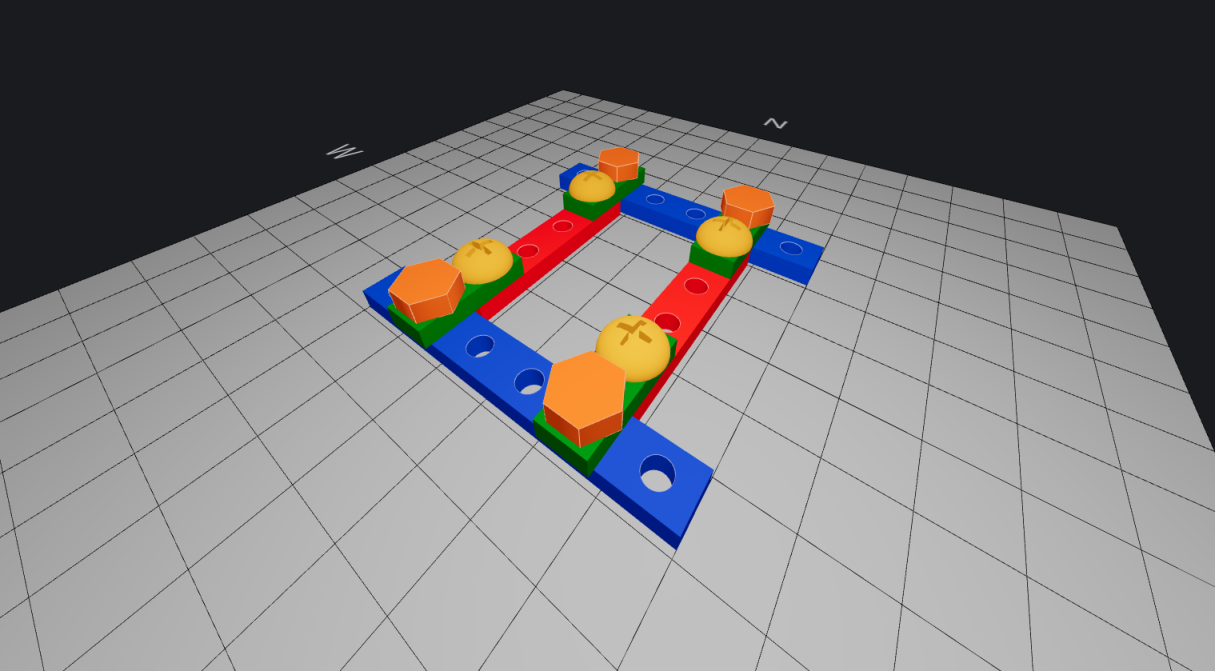}
        Cocobots
    \end{minipage}
    \caption{Graphical representations of the Minecraft and the Cocobots environments.} \label{fig:illustration}
\end{figure}

Figure~\ref{fig:illustration}  contrasts our environment with Minecraft-based tasks.

\section{Tasks 1 and 2}
\label{task-1-2-data}
\paragraph{Evaluation Protocol.}
Evaluation is performed on structured representations (part, color, coordinates, relations) rather than raw text outputs. For multi-step outputs, only the final action corresponding to each instruction is evaluated, as intermediate actions may include state reconstruction. The second instruction is classified as independent if it specifies absolute coordinates, and dependent if it refers to a previously placed part (e.g., ``on top'', ``left of''). For dependent placements, correctness is determined by relation type and anchor consistency rather than absolute coordinates. Coordinates follow a row--column convention with the top-left cell defined as (1,1).

\paragraph{Single-part placement instructions} To test Task (i), we generated 20 simple, synthetic sentences containing a placement instruction of a single object, with all the needed parameters in the input. As explained in Section~\ref{sec:coco}, the {\sc Cocobots} parts are three-dimensional, may occupy one or two cells in the grid, and are gravity-bound. Since there are no prior placed objects in these instructions, gravity constraints mean that we only specify the $[x, y]$ coordinates in these sentences. The sentence structures are the following:

    \begin{itemize}
        \item Place a \texttt{[COLOR]} \texttt{[PART]} at the \texttt{[X]}th column, \texttt{[Y]}th row.
        \item Place a \texttt{[COLOR]} \texttt{[PART]} at the \texttt{[relative coordinate]} of the board.
    \end{itemize}

\noindent For horizontal bridges, two columns are provided for the $[x_1, x_2]$ placement requirement, and, for vertical bridges, two rows for $[y_1, y_2]$. The {relative coordinates} may be grid positions such as \textit{top left},  \textit{middle}, etc., to be interpreted by the LLM agents. 

This data captures the two ways in which a user may fully define the placement of a part, either with absolute or relative coordinates. 

\paragraph{Sequences of single-part instructions} For Task (ii), the objective is to extract multiple objects from the same input. The input is composed of two sentences, with the same structure as the single instruction sentences. We created 13 tuples of these sentences with three possible variations: either the two parts are independently placed from each other, or the second part's placement is dependent on the first part's position (either on top or adjacent). 

Results of tasks 1 and 2:
We evaluated the accuracy of identifying the part type (e.g., \textit{washer}, \textit{hexagonal nut}), the color of the part, and the coordinates of the part on the grid for the placement of an object.  Table \ref{tab:task-1} shows the results.  The baselines, the agentic system, and {\sc Cocoreli}, could all identify the type and colors. However, while the agentic LLM and {\sc Cocoreli} identified correct coordinates, the CoT single LLMs struggled with this.

Our system performed well on Task (ii) as well and 

correctly identified the object properties for both objects in the input, as did the agentic model. The CoT LLMs were successful in extracting information for the first object of the input, but struggled to properly identify the second, with a 23-38\% drop in accuracy. 

For Task (ii), we evaluate the accuracy of the first and second instructions separately, since each part has a separate object and property. The results are in Table~\ref{tab:task-2}. 
 The second part of the sequence may have either independent coordinates (``Independent'') or dependent coordinates (``Dependent'') to an anchor part (``Dep. Anchor'').
\begin{table}[t]

    \centering
    \setlength{\tabcolsep}{4.4pt}
    \small
    \renewcommand{\arraystretch}{0.3}
    \begin{tabular}{@{}lcccc@{}}
    \toprule
    \textbf{Accuracy} & \textbf{GPT-4.1} & \textbf{Claude} & \textbf{Agentic} & \textbf{{\sc Cocoreli}} \\ \midrule
    {Part type} & \textbf{100} & \textbf{100} & \textbf{100} & \textbf{100} \\ \midrule
    {Part color} & \textbf{100} & \textbf{100} & \textbf{100} & \textbf{100} \\ \midrule
    {Coordinates} & 83.33 & 83.33 & \textbf{100} & \textbf{100} \\ \midrule
    {TOTAL} & 94.44 & 94.44 & \textbf{100} & \textbf{100} \\ \bottomrule
    \end{tabular}
    
    \caption{Fine-grained accuracy results (\%) on Task (i): Single-part placement. 
    }    \label{tab:task-1}

    \vspace{10pt}
    \centering
    \small
    \renewcommand{\arraystretch}{0.3}
    \setlength{\tabcolsep}{1.8pt}
    \begin{tabular}{@{}l|cc|cc|cc|cc@{}}
    \toprule
    \multirow{2}{*}{\textbf{Accuracy}} & \multicolumn{2}{c}{\textbf{GPT-4.1}} & \multicolumn{2}{c}{\textbf{Claude}} & \multicolumn{2}{c}{\textbf{Agentic}} & \multicolumn{2}{r}{\textsc{Cocoreli}} \\ \cmidrule(l){2-9} 
     & \textbf{1st} & \textbf{2nd} & \textbf{1st} & \textbf{2nd} & \textbf{1st} & \textbf{2nd} & \textbf{1st} & \textbf{2nd} \\ \midrule
    {Part type} & \textbf{100}  & \textbf{100} & \textbf{100} & \textbf{100} & \textbf{100} & \textbf{100} & {\textbf{100}} & {\textbf{100}} \\\midrule
    {Part color} & \textbf{100} & \textbf{100} & \textbf{100} & \textbf{100} & \textbf{100} & \textbf{100} & {\textbf{100}} & {\textbf{100}} \\\midrule
    {Coordinates} & \textbf{100} & -- & \textbf{100} & -- & \textbf{100} & -- & {\textbf{100}}& -- \\\midrule
    {Independent} & -- & \textbf{100} & -- & \textbf{100} & -- & \textbf{100} & -- & {\textbf{100}} \\\midrule
    {Dependent} & -- & \textbf{100} & -- & \textbf{100} & -- & \textbf{100} & -- & {\textbf{100}} \\ 
    {Dep. Anchor} & -- & 63.64 & -- & 63.64 & -- & \textbf{100} & -- & {\textbf{100}} \\ \midrule
    {TOTAL} & \multicolumn{2}{c}{96.97} & \multicolumn{2}{c}{96.97} & \multicolumn{2}{c}{\textbf{100}} & \multicolumn{2}{c}{\textbf{100}} \\ \bottomrule
    \end{tabular}
    
    \caption{Fine-grained accuracy results for Task (ii): Simple two-part instruction sequences, for the first and second instructions of the sequence. Dashes indicate not applicable (coordinates are not evaluated for dependent placements). While overall accuracy is high, performance on dependent anchor placement reveals persistent failures in executing relational instructions.}
    \label{tab:task-2}
\end{table}
{
\section{Task 3 and 5}
Here are the full results on the more complex shapes, given in Table~\ref{tab:task-3-full}
\begin{table}[t]
    \centering
    \footnotesize
    \renewcommand{\arraystretch}{0.3}
    \setlength{\tabcolsep}{5.3pt}
    \begin{tabular}{@{}lcccc@{}}  
    \toprule
    \textbf{Shape} & \textbf{GPT-4.1} & \textbf{Claude} & \textbf{Agentic} & \textbf{{\sc Cocoreli}} \\ \midrule
    {A} & 66.67 & 50.00 & \textbf{100} & \textbf{100} \\ \midrule
    {B} & 50.00 & \textbf{100} & \textbf{100} & 75.00 \\ \midrule
    {C} & \textbf{100} & \textbf{100} & \textbf{100} & \textbf{100} \\ \midrule
    {D} & 0 & 50.00 & \textbf{100} & 25.00 \\ \midrule 
    {E} & \textbf{100} & \textbf{100} & \textbf{100} & 75.00 \\ \midrule
    {G} & 60.00 & 80.00 & 0 & \textbf{100} \\ \midrule
    {X} & 0 & \textbf{100} & 50.00 & \textbf{100} \\ \midrule
    {Square} & \textbf{100} & \textbf{100} & \textbf{100} & 75.00 \\ \midrule
    {+} & \textbf{100} & \textbf{100} & \textbf{100} & \textbf{100} \\ \midrule
    {Moroccan} & 0 & 0 & 0 & 57.1 \\ \midrule
    {TOTAL} & 54.67 & 69.04 & 69.04 & \textbf{78.57} \\ \bottomrule
    \end{tabular}
    
    \caption{Accuracy (\%) based on \# correct steps\ /\#total steps following instructions in Task (iii). }
    \label{tab:task-3-full}
\end{table}

Here is the full results for the abstraction claims in Table~\ref{tab:task-v-full}
\begin{table}[t]
    \centering
    \setlength{\tabcolsep}{5pt}
    \small
    \renewcommand{\arraystretch}{0.1}
    \begin{tabular}{@{}lcccc@{}}
    \toprule
    \textbf{Shape ID} & \textbf{GPT-4.1} & \textbf{Claude} & \textbf{Agentic} & \textbf{{\sc Cocoreli}} \\ \midrule
    \textbf{A20 tower} & \textbf{T}& \textbf{T}& \textbf{T}& \textbf{T}\\ \midrule
    \textbf{C15 stack} & \textbf{T}& \textbf{T}& \textbf{T}& \textbf{T}\\ \midrule
    \textbf{D21} & \textbf{T}& \textbf{T}& \textbf{T}& \textbf{T}\\ \midrule
    \textbf{X34} & \textbf{T}& \textbf{T}& \textbf{T}& \textbf{T}\\ \midrule
    \textbf{Square} & \textbf{T}& \textbf{T}& \textbf{T} & \textbf{T}\\ \midrule
    \textbf{Triad} & \textbf{T}& \textbf{T}& F & \textbf{T}\\ \midrule
    \textbf{Face} & \textbf{T}& \textbf{T}& F & \textbf{T}\\ \midrule
    \textbf{I} & F & F & F & \textbf{T}\\ \midrule
    \textbf{Skull} & F & F & F & \textbf{T}\\ \bottomrule
    \end{tabular}
    
    \renewcommand{\arraystretch}{0.3}
    \caption{Evaluation on Task (v): Shape reproducibility from instructions for learning functions in context}
    \label{tab:task-v-full}
\end{table}
}
\subsection{{\sc Cocobots} task data}

We provide detailed data for Tasks (i) and (ii) in Appendix~\ref{task-1-2-data} and focus here on the remaining task families.

\paragraph{Complex structure construction.}
Task (iii) evaluates the ability to follow sequences of instructions describing multi-part structures. Instructions may involve standard or complex parts and can specify either absolute or relational placements (e.g., ``\textit{place a horizontal row of four purple gaskets}''). The set of evaluated shapes is listed in Table~\ref{tab:task-3}; additional examples and visualizations are provided in Appendix~\ref{app:task-iii}.

\paragraph{Underspecified instructions.}
Task (iv) evaluates handling of incomplete instructions. We construct two variants: underspecified single-sentence instructions (analogous to Task (i)) and underspecified two-sentence sequences (analogous to Task (ii)). Dataset construction details are given in Appendix~\ref{app:task-iv}.

\paragraph{Learning abstract functions in context.}
Task (v) evaluates whether models can abstract a structure from instructions and reproduce it in later interactions. Systems first construct nine complex structures from instructions and are then asked to recreate them without explicit restatement of the original steps. The evaluated structures are listed in Table~\ref{tab:task-v}, with further details in Appendix~\ref{app:task-v}.

\section{Ablation Study}
\label{Abalation}

\subsection{Controlled and implicit ablations}
We first performed small \textbf{controlled ablations} by disabling single modules on the task families where they are structurally required. 
As shown in Table~\ref{tab:micro_ablate}, each removal leads to categorical collapse (0\% or 100\% failure), indicating that these components are essential for system functionality.

\begin{table}[H]
\centering
\small
\begin{tabular}{p{2.3cm} c p{2.3cm}}
\hline
Ablation & Task subset (N) & Outcome \\
\hline
Clarification loop disabled & 20 & 0 / 20 completions \\
External memory disabled & 5 & 0 / 5 abstractions \\
Parser/Builder/Locator sep. removed & 10 & 10 / 10 invalid JSONs or crashes \\
\hline
\end{tabular}
\caption{Controlled ablation results on representative subsets of the relevant task families. Each removal produces categorical collapse (0\% or 100\% failure), confirming that these modules are structurally necessary.}
\label{tab:micro_ablate}
\end{table}

While these controlled removals establish that modules are indispensable, they do not reveal \emph{how} each contributes. 
Because {\sc Cocoreli} is prompt-driven, destructive ablations would require rewriting prompts, effectively creating a new system rather than a controlled variant. 
We therefore report \textbf{task-based ablations}, where distinct evaluation regimes (simple parsing, multi-part planning, ambiguous inputs, and abstraction with memory) naturally isolate the role of each module without altering the architecture.

\begin{itemize}
    \item \textbf{Parsing only (Tasks~i–ii).} With 1–2 parts, {\sc Cocoreli} achieves 100\% accuracy, isolating pure parsing ability (Table~\ref{tab:task-1}, Table~\ref{tab:task-2}). 
    
    \item \textbf{Complex multi-part instructions (Task~iii).} Accuracy drops as implicit planning load increases, highlighting the limits of planning without a dedicated planner (Table~\ref{tab:task-3}). 
    
    \item \textbf{Ambiguous instructions (Task~iv-a,b).} Underspecified fields produce \texttt{null} in the JSON, which deterministically triggers a clarification question (CQ). This shows the loop’s necessity: without it, the system hallucinates or stalls (Table~\ref{tab:task-iv-single}, Table~\ref{tab:task-iv-two}).

    \item \textbf{Abstraction and reuse (Task~v).} Once parsing succeeds, retrieval from external memory is deterministic, so {\sc Cocoreli} reproduces complex shapes with 100\% accuracy. Without memory, no reuse is possible (0\%), showing that memory is essential for enabling abstraction and reuse (Table~\ref{tab:task-v}).

\end{itemize}

Together, the controlled and task-based ablations show both the necessity and the functional role of parsing, clarification, and memory, without requiring ad-hoc prompt rewrites.
\section{More Related Work}
\label{sec:more-related-work}

We define an agentic system as a triple ${\cal A} = (s_0, S, \rightarrow_A)$, with $S$ a set of states, $s_0$ a special state representing the environment, and $\rightarrow_A: S \rightarrow S$ a non-empty typed set of labelled transitions $S$  with $A$ a set of action types.  Agents are functions or relations (if agents are nondeterministic) between states, while states represent: (i) inputs from the environment to an agent, (ii) a repository of information about properties and objects, possible actions, as well as goals and constraints relevant to system tasks, (iii) representations pertinent to the task that one agent provides to another agent. We can compose agents together, as the set of transitions are closed under composition $a_1 \circ a_2$, non deterministic choice $a_1 \cup a_2$, iteration $a^*$, providing general programmable relations over functions.  Our agentic definition is thus a special case of a dynamic semantics used for programming languages \cite{Harel1984}, which enables sophisticated logical techniques for studying the logical properties of agentic systems. A  

Our definition applies to an agentic system that takes a complex prompt of an LLM and divides it into a sequence of smaller prompts for different LLM agents.  We can also easily represent memory as a state that an agent may operate on by adding new transitions or adding more facts, constraints and functions.  An executor defines a transition on $s_0$.  A decision procedure is a transition that outputs a state for the executor. 

\paragraph{Prompting techniques}

LLMs falter at generalization, since it requires abstraction beyond the scope of their learned representations \citep{zhang:etal:2022}.  Approaches such as \textit{in-context learning} (ICL) provide additional information to the LLM in the form of an analogy, thus expanding their domain knowledge, for example, in mathematical equations \citep{naim:asher:2025, naim:asher:2024b, ssa, gsm-symbolic}. However, ICL remains limited to domain adaptation, lacking generalizability \citep{dong-etal-2024-survey}.

Chain-of-thought (CoT) prompting \citep{wei2022chain, wang2023selfconsistencyimproveschainthought} aims to improve performance by asking the LLM to resolve a task by documenting its solution step by step.
SayCan \citet{ahn2022icanisay} 
supply an LLM with the list of simpler tasks and prompt the model to decompose a complicated task as a sequence of simpler tasks. 
\citet{press-etal-2023-measuring} introduce multi-hop questions in CoT prompting with answers that require multiple facts that are novel to the LLM. The ReACT framework by \citet{yao2023reactsynergizingreasoningacting} extends the CoT process by integrating information retrieval from the Wikipedia API. \citet{yao2023treethoughtsdeliberateproblem} introduce trees of thought (ToT) in order to diversify the LLM's decision-making process to improve performance. \citet{long2023largelanguagemodelguided} also enhances CoT with ToT, and incorporate agents and external memories.

\paragraph{Agentic models}
Several studies have tested single-agent LLM models on their abilities in complex interactive tasks, including games, and report weaknesses in reasoning and planning \citep{liu2024agentbench, gioacchini-etal-2024-agentquest, wu2024smartplay}. Specifically on Minecraft, \citet{wu2024smartplay} report very low performance of multiple LLMs compared to human competencies and problems, such as with 3D spatial reasoning. Larger LLMs (GPT3.5, LLaMA-3.1 70B) still outperform smaller, novel LLMs in game benchmarks \citep{huang2025fardecisionmakingllmsevaluating}.  
Regarding multi-agent approaches, \citet{10.1145/3586183.3606763} introduced Generative Agents, a system based on GPT-3.5 that uses independent agents, each capable of utilizing functions for memory retrieval, reflection generation, and sequence planning. 
\citet{NEURIPS2024_5aee125f} used multi-agent reinforcement learning (MARL), testing in a variety of tasks in experimental computational game theory. 
\citet{xu-etal-2024-magic} assigned cognition, adaptability, rationality, and collaboration objectives to multiple agents, using probabilistic graphic modeling. They reported that the larger LLMs, such as GPT-4 Turbo, had the best performance on game benchmarks. 
\paragraph{Interactive and memory-augmented agents}
\citet{mehta2024improvinggroundedlanguageunderstanding} study interaction in Minecraft-style environments, where agents can receive help feedback from humans or simulated sources. Their approach is reactive and correction-oriented: the agent acts, and when errors occur, feedback is integrated to recover. In contrast, the {\sc Cocobots} benchmark assumes that wrong placements are not permissible, making proactive clarification essential. {\sc Cocoreli}’s discourse module therefore resolves underspecification before execution, whereas Mehta’s framework is not directly applicable under {\sc Cocobots}’s stricter setting.

\citet{sarch2023openendedinstructableembodiedagents} introduce HELPER, an embodied agent with a memory-augmented LLM that retrieves past language–program pairs to improve generalization in instruction following. While effective on TEACh, HELPER’s reliance on retrieval limits its ability to handle novel underspecified tasks, as it lacks the compositional abstractions (e.g., coordinates, parts, colors) needed for recomposition in {\sc Cocobots}. Moreover, HELPER’s feedback mechanism is reactive and post-execution, asking for corrections after acting, whereas {\sc Cocobots} requires proactive clarification before any placement, since errors are not permissible. Consequently, HELPER’s memory and correction style are not directly applicable to the structured underspecification benchmarks that {\sc Cocoreli} addresses.

\paragraph{Tool learning}
Tool learning involves the use of external tools to improve problem solving and broaden the functional scope of the LLM, thus augmenting performance on specific downstream tasks \citep{parisi2022talmtoolaugmentedlanguage, pmlr-v202-gao23f, long2023largelanguagemodelguided}.  Recently, researchers have developed self-supervised tools that can support a level of abstraction and generalization, making them useful across multiple applications.
\citet{schick2023toolformer} developed Toolformer, an LLM with GPT-3 in its underlying architecture, that is capable of generating functions to use API tools from various websites. 
ToolLLM \citep{qin2023toolllmfacilitatinglargelanguage} has a similar motivation in generating API calling functions with LLaMA-2 7B in an unsupervised manner, in addition to proposing ToolBench, a dataset for evaluating unsupervised tool learning. Gorilla \citep{NEURIPS2024_e4c61f57} is also an API function generating tool that incorporates document retrieval from the target API documentation.  
\citet{tang2023toolalpacageneralizedtoollearning} introduced ToolAlpaca, a framework to fine-tune LLMs for better tool learning of API functions. 

\section{Cost Analysis}
\label{cost-analysis}
Table~\ref{tab:efficiency} compares efficiency across different setups for instruction-to-action parsing. 

The \textbf{CoT systems} (GPT-4.1, Claude) incur direct API costs per query. Their outputs are compact (around 20–40 tokens), which keeps costs modest, but differences in vendor pricing lead to GPT averaging around \$6 per 1K queries and Claude around \$10. Latency is flat (8–11s) across specified and underspecified tasks because these models do not attempt clarification—underspecified inputs are simply passed through with \texttt{null} placeholders. This avoids additional cost but leaves instructions incomplete.

The \textbf{Agentic baseline} (LLaMA-70B) uses multi-step reasoning with longer outputs (around 150–300 tokens). When underspecified, it may issue clarification questions, which slightly increases output length and cost, but this behavior is inconsistent: some queries are left unresolved while others expand into long reasoning traces. As a result, the underspecification penalty manifests as stochastic increases in token usage, with the burden of resolution falling on the model’s exploratory generation.

\textbf{{\sc Cocoreli}} (LLaMA-8B) runs locally and therefore incurs \emph{no} API costs. Its outputs are structured JSON parses, where missing fields are explicitly marked as \texttt{null}. A deterministic Python layer \textbf{executor module} then converts these into targeted clarification questions. This design shifts underspecification handling out of the LLM itself: the model cost remains constant across specified and underspecified tasks, and clarifications are resolved predictably without bloating output sequences.

In short, GPT and Claude bypass underspecification, Agentic absorbs it stochastically in model outputs, and {\sc Cocoreli} handles it deterministically outside the model, trading modest latency for zero API cost and predictable clarification.

We do not report efficiency for the abstraction experiments in Table \ref{tab:task-v}, since these are one-shot queries rather than repeated instruction following. 
Nonetheless, the scaling behavior is informative: 
for \textbf{CoT systems}, output size grows directly with the number of parts (e.g., regenerating 60 \texttt{place(...)} calls is much more expensive than 6). 
For the \textbf{Agentic baseline}, outputs scale with the complexity of the induced program (longer Python code for more complex structures). 
By contrast, \textbf{{\sc Cocoreli}} abstracts structures into a fixed-size JSON schema, so output size is independent of the structure’s size or complexity. 
This makes {\sc Cocoreli}’s abstraction mechanism cost-invariant, even as structures grow large.

\hidden{

\subsection{{\sc Cocobots} task data}

We provide detailed data for Tasks (i) and (ii) in Appendix~B and focus here on the remaining task families.

\paragraph{Complex structure construction.}
Task (iii) evaluates the ability to follow sequences of instructions describing multi-part structures. Instructions may involve standard or complex parts and can specify either absolute or relational placements (e.g., ``\textit{place a horizontal row of four purple gaskets}''). The set of evaluated shapes is listed in Table~\ref{tab:task-3}; additional examples and visualizations are provided in Appendix~\ref{app:task-iii}.

\paragraph{Underspecified instructions.}
Task (iv) evaluates handling of incomplete instructions. We construct two variants: underspecified single-sentence instructions (analogous to Task (i)) and underspecified two-sentence sequences (analogous to Task (ii)). Dataset construction details are given in Appendix~\ref{app:task-iv}.

\paragraph{Learning abstract functions in context.}
Task (v) evaluates whether models can abstract a structure from instructions and reproduce it in later interactions. Systems first construct nine complex structures from instructions and are then asked to recreate them without explicit restatement of the original steps. The evaluated structures are listed in Table~\ref{tab:task-v}, with further details in Appendix~\ref{app:task-v}.
}
\begin{table}[t]
\centering
\small
\setlength{\tabcolsep}{2.5pt}
\renewcommand{\arraystretch}{0.9}
\begin{tabularx}{\columnwidth}{l l c c c}
\hline
\textbf{System} & \textbf{Task} & \textbf{Tokens} & \textbf{Latency(s)} & \textbf{Cost(\$/1K)} \\
\hline
\textbf{GPT} & 1S & 2818/17 [2835] & 10 & 5.77 \\
                 & 2S & 2829/33 [2862] & 10 & 5.92 \\
                 & 1U & 2816/17 [2833] & 11 & 5.77 \\
                 & 2U & 2829/33 [2862] & 10 & 5.92 \\
\midrule
\textbf{Claude}  & 1S & 3096/19 [3115] & 8 & 9.57 \\
                 & 2S & 3139/38 [3177] & 8 & 9.99 \\
                 & 1U & 3126/19 [3145] & 8 & 9.66 \\
                 & 2U & 3139/38 [3177] & 8 & 9.99 \\
\midrule
\textbf{Agentic} & 1S & 2823/150 [2973] & 9 & 2.79 \\
                   & 2S & 2835/299 [3134] & 9 & 2.94 \\
                   & 1U & 2823/150 [2973] & 9 & 2.79 \\
                   & 2U & 2835/299 [3134] & 9 & 2.94 \\
\midrule
\textbf{{\sc Cocoreli}} & 1S & 4623/141 [4764] & 12 & 0.00 \\
                   & 2S & 4635/282 [4917] & 12 & 0.00 \\
                   & 1U & 4623/141 [4764] & 10 & 0.00 \\
                   & 2U & 4635/282 [4917] & 12 & 0.00 \\
\hline
\end{tabularx}
\caption{Efficiency metrics across systems and task types. 
Tokens are average per-instruction values, reported as input/output [total]. 
Costs are reported in U.S. dollars per 1,000 queries, computed from official per-million input/output token pricing (\textsc{Cocoreli} runs locally, cost $0.00$). 
1S = one-part specified, 2S = two-part specified, 1U = one-part underspecified, 2U = two-part underspecified. 
Task~3 omitted since costs scale proportionally with instruction length while relative ordering is preserved (accuracy in Table~\ref{tab:task-3}).}

\label{tab:efficiency}
\end{table}

\section{User Burden Analysis}
\label{user-burden}
To evaluate the practical impact of our system on end users, we conduct a \emph{user burden study} that quantifies both the correctness and the effort required during interaction. 
In the context of our \textsc{Cocobots} setup, user burden primarily arises from the clarifying questions (CQs) that the system issues to resolve underspecified instructions. 

\subsection{Metrics for Underspecified Instructions}

For instructions that are partially underspecified, we define the following metrics with respect to the gold standard set of missing information:

\begin{itemize}
    \item \textbf{True Positive (TP) CQs}: A CQ that correctly targets missing information. This reflects \emph{necessary questions} that genuinely reduce ambiguity for the user.
    
    \item \textbf{False Positive (FP) CQs}: A CQ that asks for information which is already fully specified. These represent \emph{unnecessary interruptions}, increasing cognitive load and interaction time.
    
    \item \textbf{False Negative (FN) CQs}: Missing information that the system fails to ask about. FN CQs indicate situations where the user must provide additional guidance on their own.
    
    \item \textbf{Precision, Recall, F1}: Derived from TP, FP, and FN. Precision reflects the fraction of CQs that were necessary (low FP), recall measures the fraction of missing information addressed (low FN), and F1 captures the balance. These metrics quantify \emph{user burden in terms of correctness}.
\end{itemize}
By combining correctness (or user-burden) metrics (TP, FP, FN, precision, recall, F1 for underspecified instructions) with unnecessary CQs (for fully specified instructions) and efficiency measures (tokens, time, cost), we provide a comprehensive and objective assessment of user burden. 

These metrics provide a reproducible proxy for user burden. While actual human perception may vary, unnecessary and missed clarifications, as well as time and token usage, directly reflect potential cognitive and operational load on the user.

\section{{\sc Cocobots} dataset details}



\subsection{Task (iii)}\label{app:task-iii}

For this task, every single instruction is fully specified. The instructions themselves are complex, drawing away from the simple template style approach we were following.  The shapes are constructed from 2 to 6 instructions.  Some of these contain the same type of part, while others have a variety of parts and colors. The Moroccan Bridge structure is a highly complex shape composed of 18 {\sc Cocobots} parts.  A detailed list and a visualization of the shapes in the {\sc Cocobots} for this task is below. 
 \noindent Here is a sample:
\begin{quote}
    \textbf{[Turn 1]}<Architect>Starting on the second row and second column, place a magenta washer\\
    \textbf{[Turn 2]}<Architect> On the third row, a column of four more magenta washers. All in the second column\\ \textbf{...}
\end{quote}

\begin{figure}[!h]
    \centering
    \includegraphics[width=0.45\textwidth]{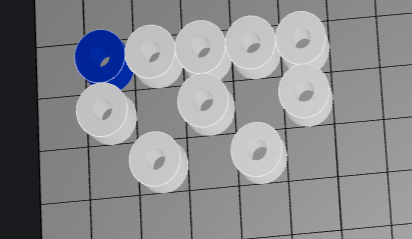}
    \caption{The B Shape}\label{fig:b-shape}
\vspace{20pt}
    \centering
    \includegraphics[width=0.45\textwidth]{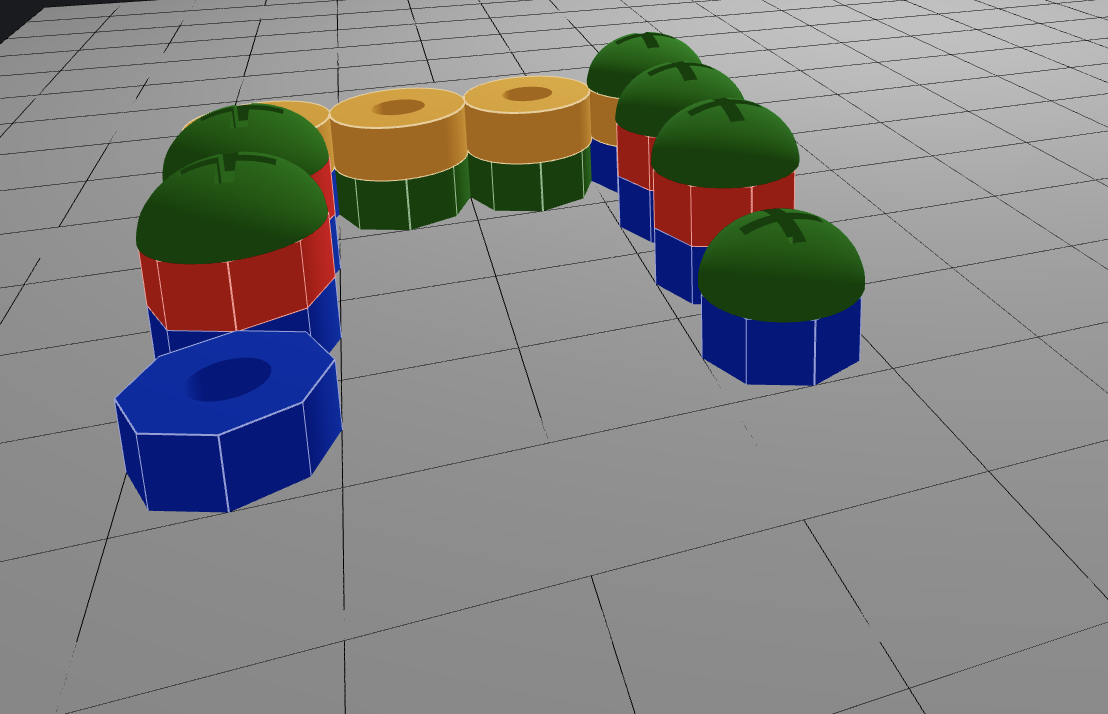}
    \caption{The Moroccan Bridge}\label{fig:moroccan-shape}
\end{figure}

\noindent A detailed list of the 10 complex shapes used for Task (iii): Construction of complex shapes \\
\begin{itemize}
    \item \textbf{A}: 14 parts (magenta washers) | 6 instructions
    \item \textbf{B}: 10 parts (various parts/colors) | 4 instructions (see Figure~\ref{fig:b-shape})
    \item \textbf{C}: 8 parts (green screws) | 3 instructions
    \item \textbf{D}: 10 parts (various parts/colors) | 4 instructions
    \item \textbf{E}: 10 parts (various parts/colors) | 4 instructions
    \item \textbf{G}: 17 parts (various parts/colors) | 5 instructions 
    \item \textbf{+}: 9 parts (magenta bolts) | 3 instructions
    \item \textbf{Square}: 16 parts (various bolts) | 4 instructions
    \item \textbf{X}: 10 parts (various bolts) | 2 instructions
    \item \textbf{Moroccan Bridge structure}: A composite shape, made up of a complex shape that has to be learned as a \textbf{Moroccan Bridge} (5 parts of various colors) and reused, alongside 5 other standard parts | 8 instructions (see Figure~\ref{fig:moroccan-shape}).
\end{itemize}

\hidden{
\begin{table}[t]
    \centering
    \footnotesize
    \renewcommand{\arraystretch}{0.3}
    \setlength{\tabcolsep}{5.3pt}
    \begin{tabular}{@{}lcccc@{}}  
    \toprule
    \textbf{Shape} & \textbf{GPT-4.1} & \textbf{Claude} & \textbf{Agentic} & \textbf{{\sc Cocoreli}} \\ \midrule
    {A} & 66.67 & 50.00 & \textbf{100} & \textbf{100} \\ \midrule
    {B} & 50.00 & \textbf{100} & \textbf{100} & 75.00 \\ \midrule
    {C} & \textbf{100} & \textbf{100} & \textbf{100} & \textbf{100} \\ \midrule
    {D} & 0 & 50.00 & \textbf{100} & 25.00 \\ \midrule 
    {E} & \textbf{100} & \textbf{100} & \textbf{100} & 75.00 \\ \midrule
    {G} & 60.00 & 80.00 & 0 & \textbf{100} \\ \midrule
    {X} & 0 & \textbf{100} & 50.00 & \textbf{100} \\ \midrule
    {Square} & \textbf{100} & \textbf{100} & \textbf{100} & 75.00 \\ \midrule
    {+} & \textbf{100} & \textbf{100} & \textbf{100} & \textbf{100} \\ \midrule
    {Moroccan} & 0 & 0 & 0 & 57.1 \\ \midrule
    {TOTAL} & 54.67 & 69.04 & 69.04 & \textbf{78.57} \\ \bottomrule
    \end{tabular}
    
    \caption{Accuracy (\%) on based on \# correct steps\ /\#total steps following instructions in Task (iii). }
    \label{tab:task-3}
\end{table}
}

\subsection{Task (iv)}\label{app:task-iv}
We detail here the two data sets of underspecified instructions.
For the single sentence dataset, we created 81 sentences, similar to those of Task (i), but with one piece of information missing: the part name (e.g. ``it'' instead of ``nut''), the color, or one or all of the coordinates. 
For two sentence dataset, we created 202 instructions, in which either one or both parts described in the instruction were underspecified: the first single part placement was fully specified in 40 of those, and the second part placement was fully specified in 73 of the instructions. 

\subsection{Task (v)}\label{app:task-v}

We give details on the 9 structures for this task.  The first 4 structures are composed of 2-3 parts and have novel names (e.g., \textit{A20 tower}) in order to ensure that the LLMs did not retrieve shape information from representations learned in pretraining. The other 5 structures have common names (e.g., \textit{square}, \textit{skull}) but are composed of significantly more parts (16-18 for moderate complexity, up to 62 for complex structures).

\noindent A \textbf{sample} of human input to {\sc Cocoreli} for the C15 shape:

\begin{quote}
    \begin{itemize}
    \item[-] {INST:} Can you place a blue screw at row 4 column 5 height 1
    \item[-] {A:} Place a blue screw at row 4 column 5 height 1
    \item[-] {INST:} Place a red screw next to the blue screw, and put a red screw on top.
    \item[-] {A:} Place a red screw at row 4 column 6 height 1
    \item[-] {A:} Place a red screw at row 4 column 6 height 2
    \item[-] {INST:} This is what I call a C15 
    \item[-] {INST:} Now make me another C15 at the eighth row and ninth column 
    \end{itemize}
\end{quote}

\section{Algorithms for Abstraction and Reconstruction}

\begin{algorithm}[H]
\small
\caption{Convert Structure to Graph to \qquad Abstract Coordinates}\label{alg:cap}
\begin{algorithmic}
\Require S \Comment{\textbf{S}:Structure containing exact coordinates and part names}\\
C $\gets$ {Number of connected parts in S}\\
A $\gets$ 1 \Comment{Initialise to the first connected component}\\
G $\gets$ [ ] \Comment{Empty Adjacency List for a single component}\\
T $\gets$ \{\} \Comment{Empty Dict to store all Adjacency Lists}
\While{$C \neq 0$} \Comment{Iterate through all connected parts}
\While {v in A} \Comment{Iterate through the vertices in A}
\State $X \gets$ parts in all adjacent coordinates of v
\State G[v] $\gets$ X \Comment{Vertex v stores information about all the 'neighbors'}
\State v $\gets$ v + 1 \Comment{Move to the next vertex}

\EndWhile
\State $T[C] \gets $ G \Comment{Add G to the final graph}
\State $G \gets$ 0 \Comment{Flush G to store next component}
\State $C \gets$ $C-1$ \Comment{Move to the next component}
\EndWhile
\end{algorithmic}
\end{algorithm}

\begin{algorithm}[H]
\small
\caption{Compute Offsets from Reference Point}
\begin{algorithmic}
\Require $S_t$ \Comment{The user start coordinates for the first component}
\Require $T$ \Comment{The Stored Graph for a structure}
\State Firsts $\gets$ First vertices for each component
\State Ref $\gets$ Firsts[0] \Comment{Store the first point as reference}
\State offsets $\gets []$ \Comment{Initialize empty list to store offsets}
\State Starts $\gets$ [] \Comment{Initialize empty list to store start points}
    \For{each $(x, y, z)$ in Firsts}
        \State Append $(x - Ref.x, y - Ref.y, z - Ref.z)$ to offsets
    \EndFor
\For{each $(dx, dy, dz)$ in offsets}
        \State Append $(S_t.X + dx, S_t.Y + dy, S_t.Z + dz)$ to Starts
    \EndFor
\end{algorithmic}
\Return Starts \Comment{Contains the new starts for each component}
\end{algorithm}

\begin{algorithm}[H]
\small
    \caption{Apply a stored structure at New user Location}
\begin{algorithmic}
    \Require Starts \Comment{The start coordinates to replicate a structure}
    \Require T \Comment{The graph stored for the the required structure}\\
    C $\gets$ {The list of connected components in T}\\
    C$_{Starts}$ $\gets$ The start coordinates for each coordinate from (2)
    i $\gets$ 0 \Comment{initialize to first component index}
    \While{i $<$ len(C)} \Comment{Iterate through the components}\\
        s $\gets$ C$_{Starts}$[i] \Comment{Take the first component's start}\\
        BFS(C[i],s) \Comment{Do a bfs traversal starting from s to generate co-ords respective to the start s}\\
        i $\gets$ i + 1
    \EndWhile
\end{algorithmic}
\end{algorithm}

\begin{algorithm}[H]
\small
    \caption{Scale a shape to certain coordinates} \label{alg:scaling}
    \begin{algorithmic}
    \Require G \Comment{The concrete coordinates as they are placed on the board}
    \Require $T_x,T_y,T_z$ \Comment{The target coordinates to scale to}
    \State $S_x,S_y,S_z \gets$ The bounding box for G \\
    $S_G \gets$ Nearest\_Neighbor([$S_x,S_y,S_z$],[$T_x,T_y,T_z$])
    \end{algorithmic}
    \Return $S_G$ \Comment{Return the new coordinates after scaling to the larger grid}
\end{algorithm}

\begin{algorithm}[H]
\small
    \caption{Abstract2}
    \begin{algorithmic}
        \Require T \Comment{The relative coordinates from (1)}\\
        \Comment{These values below come from the Locator and Chatter agents}
        \State C $\gets$ Y/N \Comment{Yes/No whether to abstract color}
        \State P $\gets$ Y/N \Comment{Yes/No whether to abstract parts}
        \State S $\gets$ Y/N \Comment{Yes/No whether to abstract shape}
    \end{algorithmic}
    \Return C,P,S \Comment{Which parameters of a graph to abstract over}
\end{algorithm}

\begin{algorithm}[H]
\small
    \caption{Apply2}
    \begin{algorithmic}
        \Require G \Comment{Applied Graph from (3)}
        \Require C,P,S \Comment{Output from (5) for graph G}
       
        \If {$C \gets Y$ and $P \gets Y$ and $S \gets Y$}
             \Require $C_{val} \gets $ {From Chatter(new color)}
             \Require $P_{val} \gets $ {From Chatter and Locator(new Part)}
             \Require $S_{val} \gets $  {From Chatter(new Dimension)}\\
            G' $\gets$  Algo(3) with $C_{val},P_{val}$\\
            Final $\gets$ Algo(4) on G' with $S_{val}$
        \EndIf 
        \If {$C \gets N$ and $P \gets Y$ and $S \gets Y$}
             \Require $P_{val} \gets $ {From Chatter and Locator(new Part)}
             \Require $S_{val} \gets $  {From Chatter(new Dimension)}\\
            G' $\gets$  Algo(3) with $P_{val}$\\
            Final $\gets$ Algo(4) on G' with $S_{val}$
        \EndIf
        \If {$C \gets N$ and $P \gets N$ and $S \gets Y$}
             \Require $S_{val} \gets $  {From Chatter(new Dimension)}\\
            G' $\gets$  Algo(3) with $P_{val}$\\
            Final $\gets$ Algo(4) on G' with $S_{val}$
        \EndIf
    \end{algorithmic}
\end{algorithm}

Algorithm~\ref{alg:scaling} adapts the nearest neighbor scaling algorithm \citep{1168143}, so it includes the limitations that are native to the main algorithm.

\section{Data preparation for Toolbench}\label{sec:toolbench-data}
\hidden{
\paragraph{Data} 
We first extracted 100 workflows from the dataset, i.e., a sequence of user instructions, actions, and one or more API functions to fulfill the user's request. We ensured that they include a descriptive user instruction and an associated API function with at least two variables (non-boolean), in order to evaluate the abstractive abilities of our systems. For example, the user input ``\textit{Schedule a meeting with Alex tomorrow at 3 pm and send a confirmation email.}'' leads to the functions \texttt{create\_event(q, time)} and \texttt{send\_email(q, message)}. These functions are abstract and can be reused with different variables. 
A sample datapoint looks like the following example:-
\begin{quote}
\textbf{USER:} Schedule a meeting with Alex tomorrow at 3 pm and send a confirmation email. \\
\textbf{ACTIONS:} \texttt{create\_event(q: ``Alex'', time: ``2025-05-09 15:00'')} \\
\textbf{ACTIONS:} \texttt{send\_email(q: ``Alex'', message: ``Meeting Confirmation'', ``See you at 3 pm tomorrow!'')} \\
\textbf{USER:} This is \texttt{workflow\_1}. \\
\textbf{USER:} Please apply \texttt{workflow\_1} with query: \texttt{q:Jordan, time:2025-05-12 14:00}.
\end{quote}

For example, the instruction
\textit{"Schedule a meeting with Alex tomorrow at 3 pm and send a confirmation email."}
is represented by the workflow
\texttt{create\_event(q,time)} $\rightarrow$
\texttt{send\_email(q,message)}.
The stored workflow is later reused with new variable assignments.
}

\paragraph{Data}
We first extracted 100 workflows from the dataset, i.e., a sequence of user instructions, actions, and one or more API functions to fulfill the user's request. We ensured that they include a descriptive user instruction and an associated API function with at least two variables (non-boolean), in order to evaluate the abstractive abilities of our systems.

For example, the instruction
\textit{"Schedule a meeting with Alex tomorrow at 3 pm and send a confirmation email."}
is abstracted into the workflow
\texttt{create\_event(q,time)} $\rightarrow$
\texttt{send\_email(q,message)},
which is stored in memory as \texttt{workflow\_1}. At inference time, the stored workflow can be reused with new variables, e.g.,
\texttt{q: Jordan, time: 2025-05-12 14:00},
without relearning the workflow structure.

\noindent We present a detailed list of the 9 complex shapes used for Task (v): Abstractive instructions. The instructions for these shapes are stored in the Dialogue History.  
\hidden{
\begin{itemize}
    \item \textbf{A20 tower}: 4 parts (various) | 5 instructions
    \item \textbf{C15 stack}: 7 parts (various) | 6 instructions (see Figure~\ref{fig:c15-shape})
    \item \textbf{D21}: 6 parts (various) | 5 instructions
    \item \textbf{X34}: 5 parts (various) | 5 instructions
    \item \textbf{Square}: 16 parts (blue washers) | 17 instructions
    \item \textbf{Triad}: 18 parts (red nuts) | 17 instructions
    \item \textbf{Face}: 47 parts (blue nuts) 
    \item \textbf{I}: 18 parts (blue nuts) (see Figure~\ref{fig:I_original-shape})
    \item \textbf{Skull}: 62 parts (blue nuts) | 63 instructions (see Figure~\ref{fig:skull-shape})
\end{itemize}
}
The nine structures used in Task (v) are: \textbf{A20 tower} (4 parts, 5 instructions), \textbf{C15 stack} (7 parts, 6 instructions; ), \textbf{D21} (6 parts, 5 instructions), \textbf{X34} (5 parts, 5 instructions), \textbf{Square} (16 parts, 17 instructions), \textbf{Triad} (18 parts, 17 instructions), \textbf{Face} (47 parts), \textbf{I} (18 parts;), and \textbf{Skull} (62 parts, 63 instructions; Figure~\ref{fig:failure_cases}).
\vspace{20pt}

\hidden{
\begin{figure*}[h]
    \centering
    \includegraphics[width=0.45\textwidth]{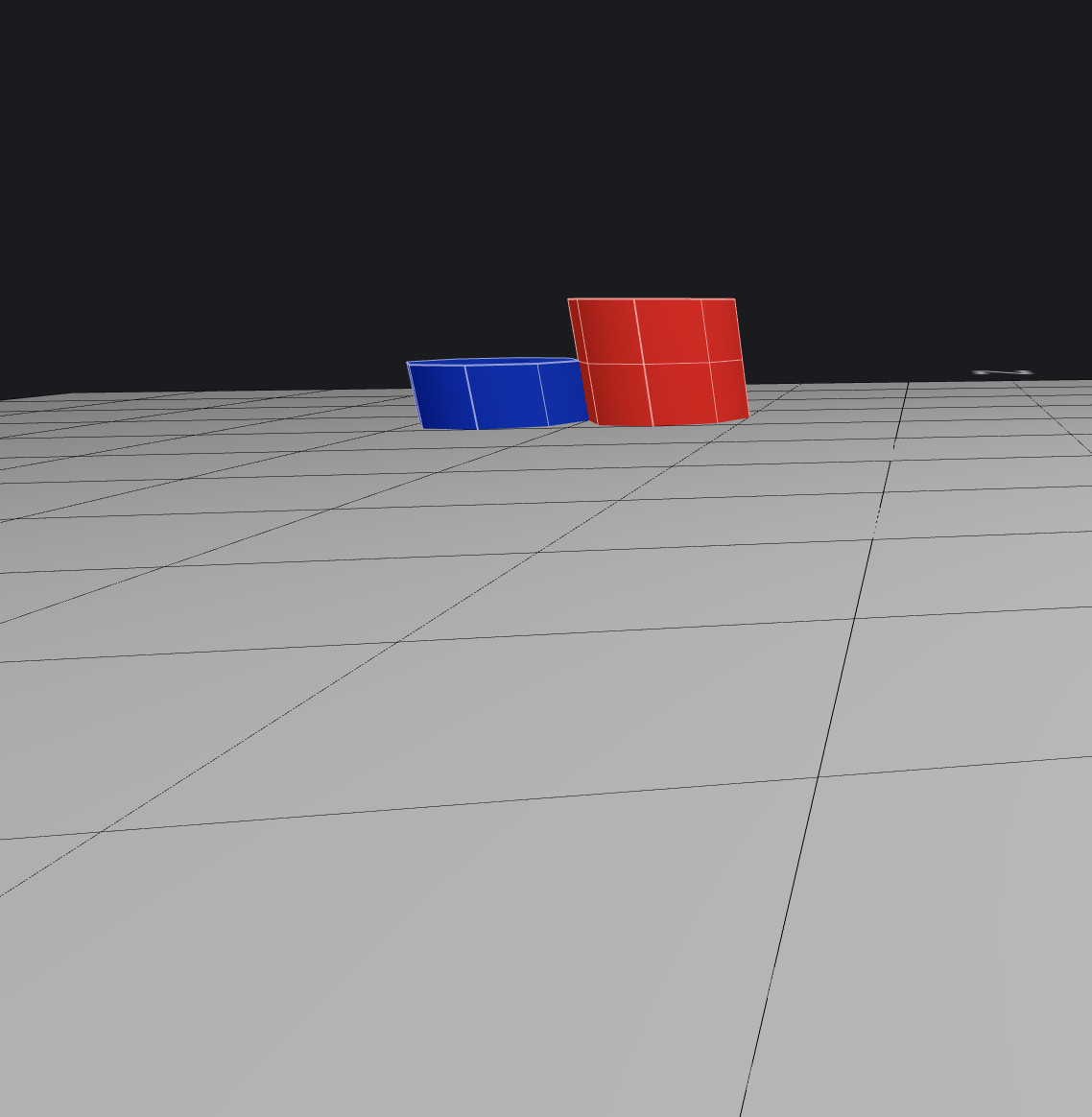}
    \caption{The C15 shape}\label{fig:c15-shape}
\vspace{20pt}
    \includegraphics[width=0.45\textwidth]{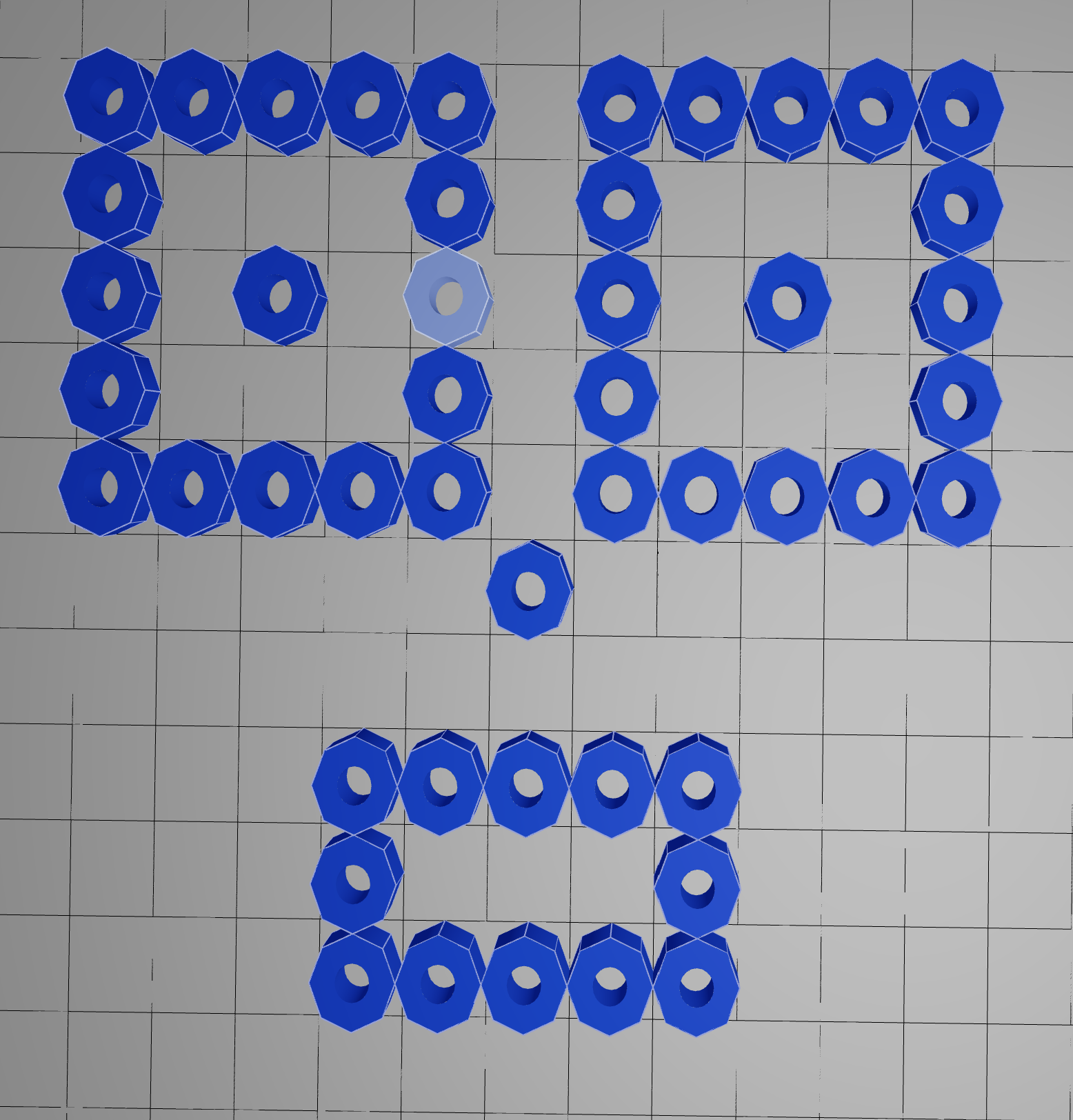}
    \caption{The Face shape}
    \label{fig:face-shape}
\vspace{5pt}
\end{figure*}

\begin{figure*}[t]
    \centering
    \includegraphics[width=0.45\textwidth]{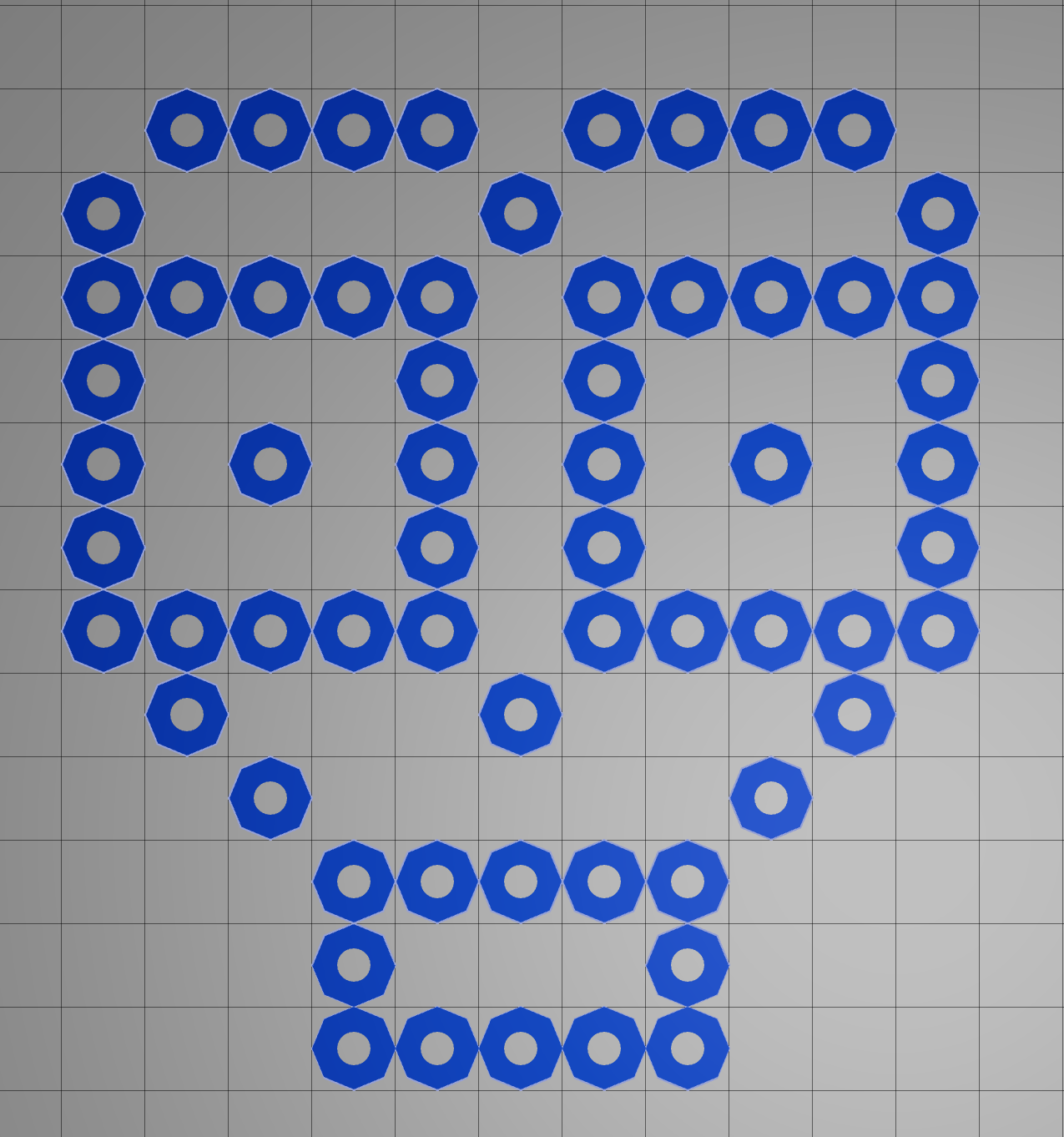}
    \caption{The Skull shape}\label{fig:skull-shape}
\end{figure*}
}
\hidden{
\begin{figure*}[t]
\centering

\includegraphics[width=0.31\textwidth]{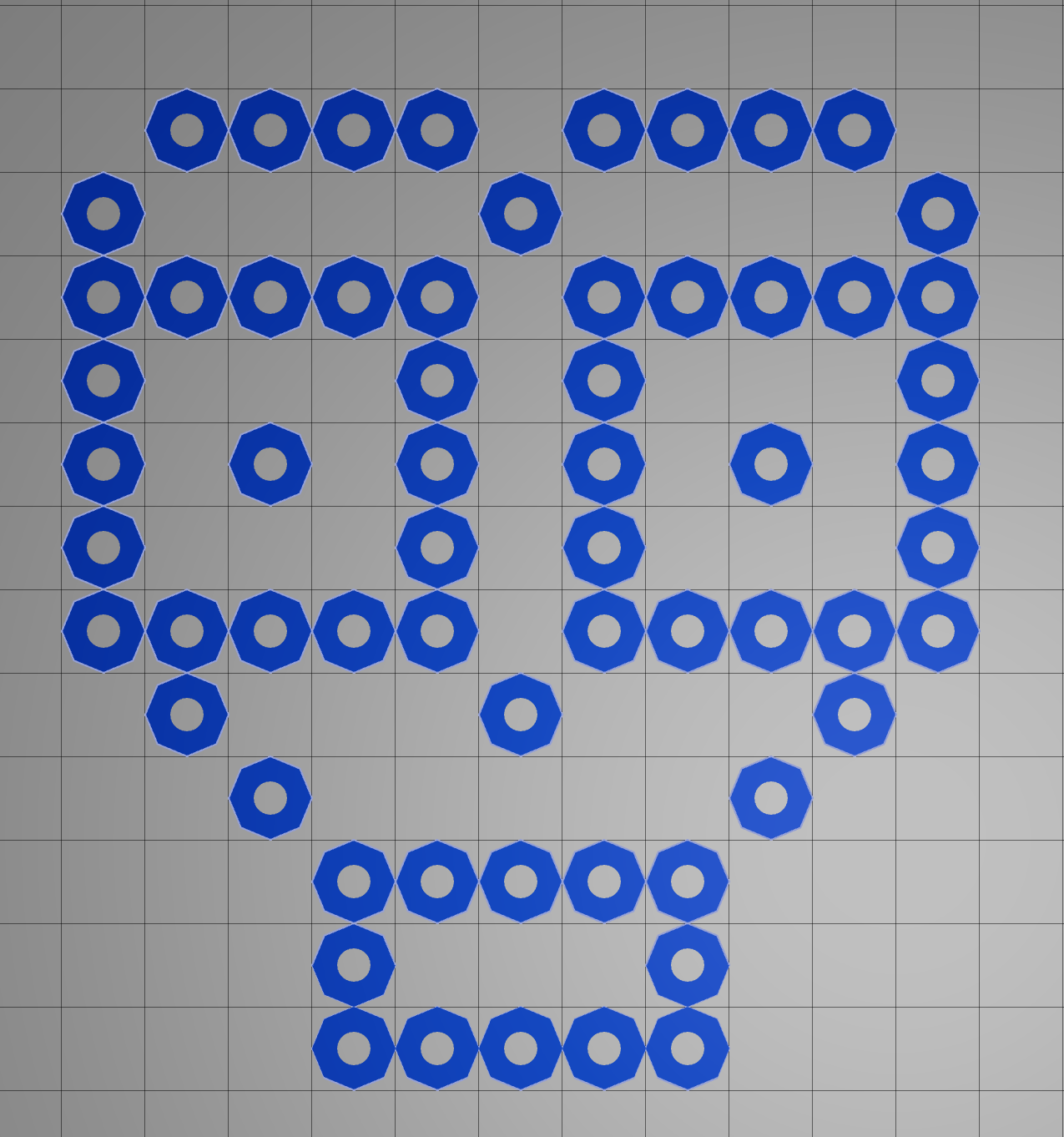}

\caption{Representative shapes used in Task (v): Skull (right).}
\label{fig:task5-shapes}

\end{figure*}
}
\section{Failure cases of the models}
\begin{figure}[t]
\centering

\includegraphics[width=0.42\columnwidth]{latex/skull.png}\hfill
\includegraphics[width=0.42\columnwidth]{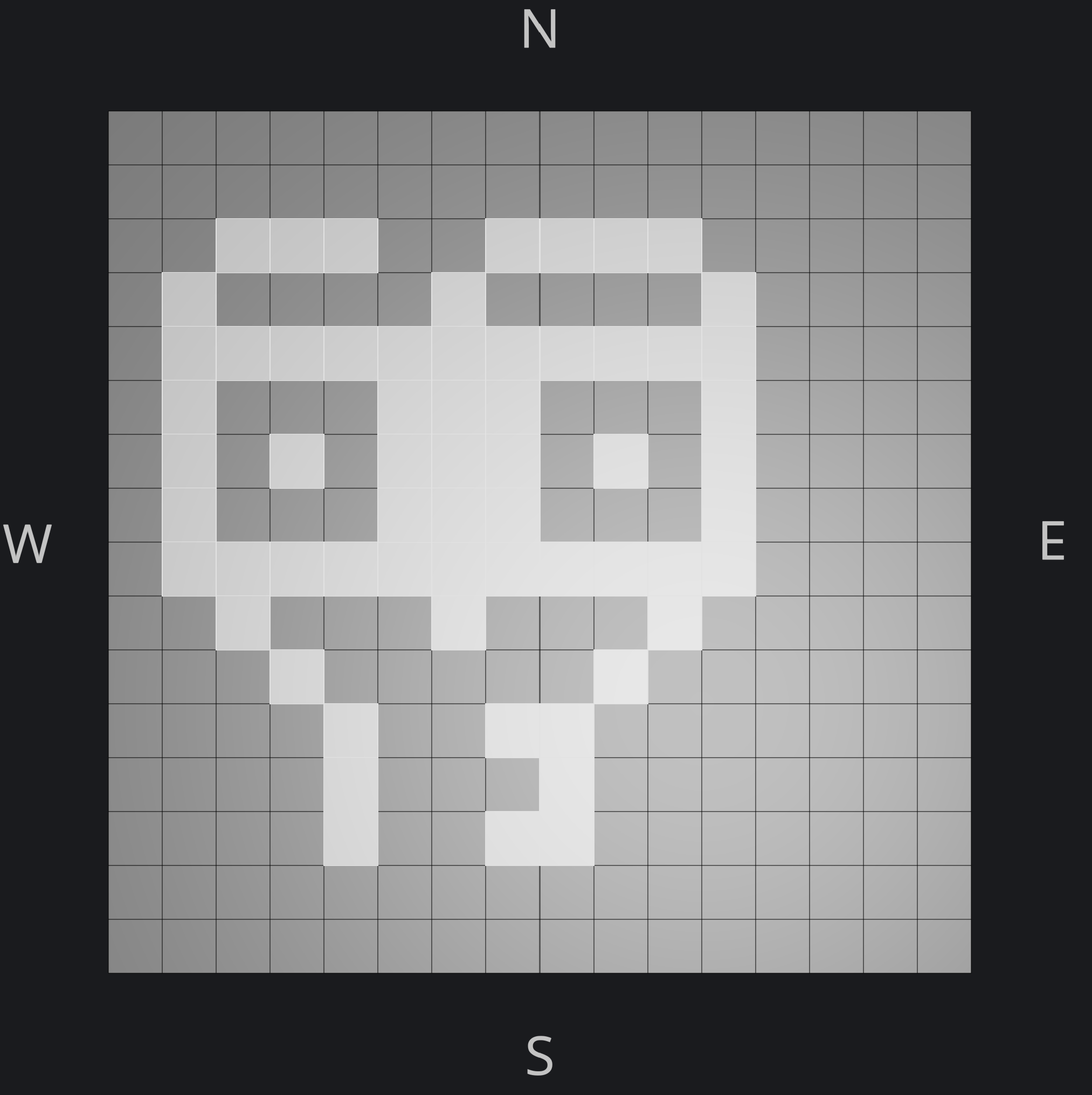}

\vspace{1mm}

\includegraphics[width=0.42\columnwidth]{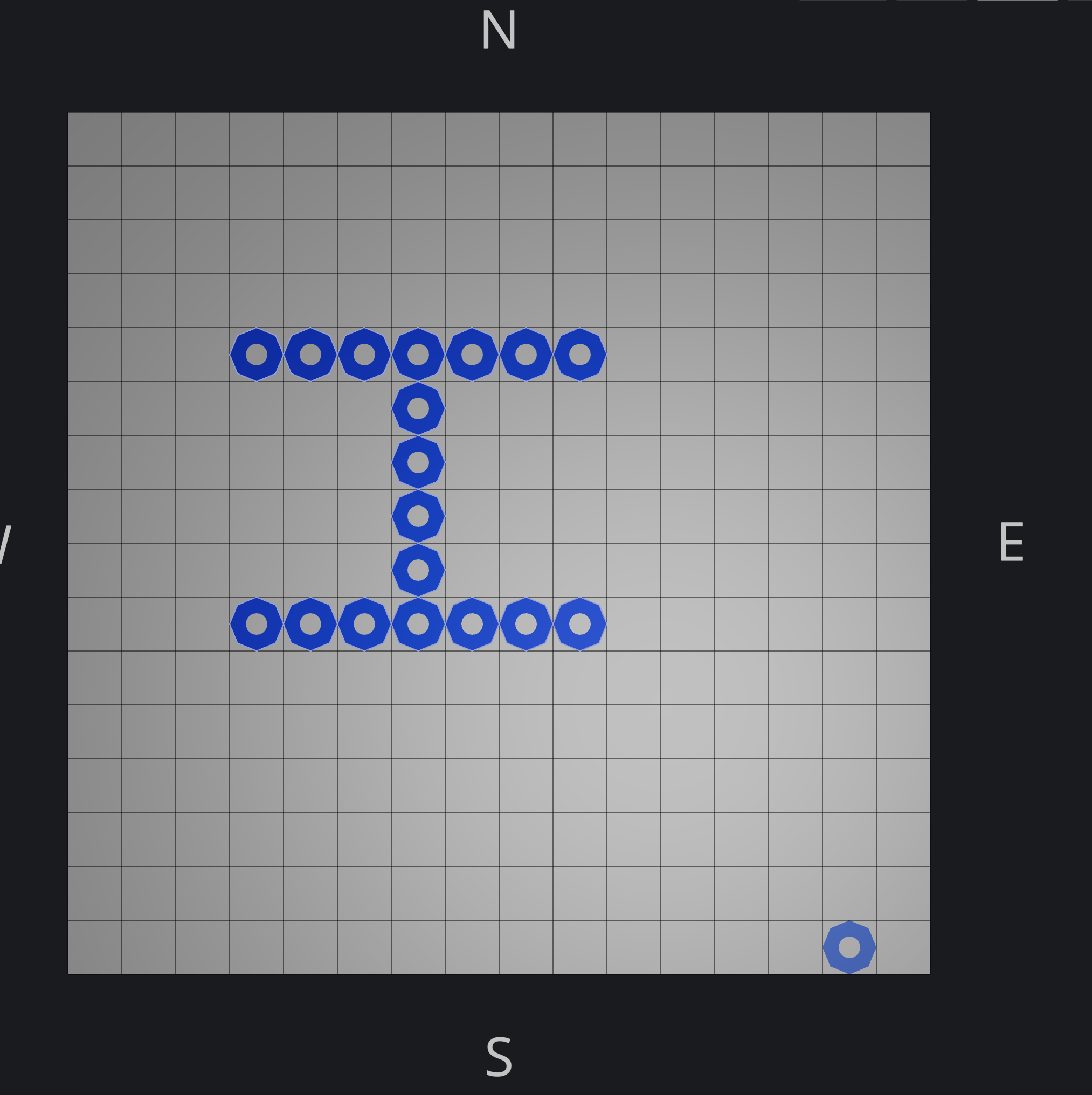}\hfill
\includegraphics[width=0.42\columnwidth]{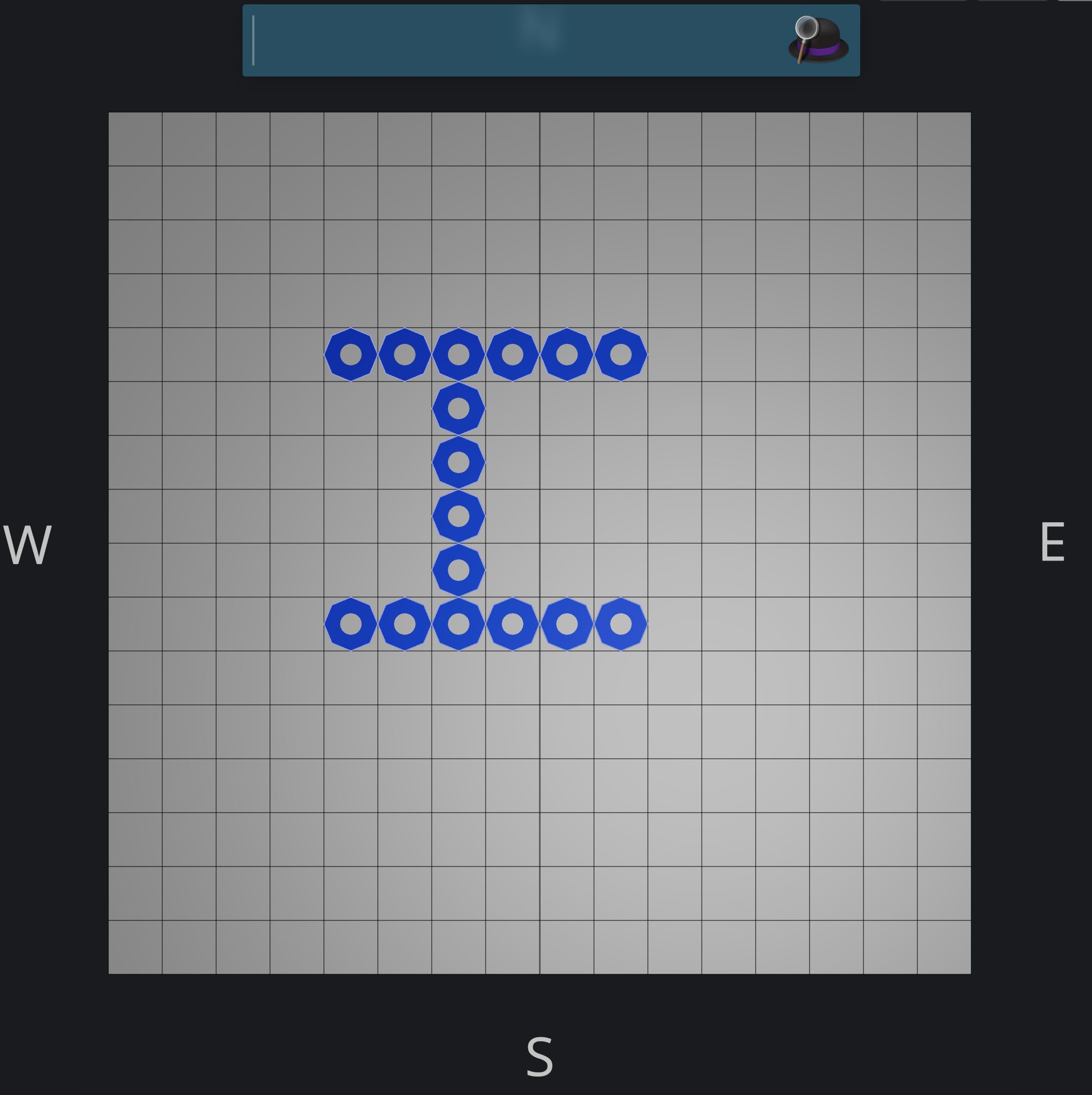}

\caption{Representative failure cases on Task~(v). Top: target Skull shape (left) and GPT-4.1 reconstruction (right). Bottom: target I shape (left) and GPT-4.1 reconstruction (right). Higher resolution versions are available in the repository.}
\label{fig:failure_cases}
\end{figure}

We present two failure cases on Task (v) from the GPT-4.1 baseline in Figure~\ref{fig:failure_cases}. As discussed in Section~\ref{sec:results} and shown in Table~\ref{tab:task-v}, the baseline was not able to reproduce the more complex shapes of the task from the dialogue history, but {\sc Cocoreli} was, through abstraction. 

\end{document}